\documentclass{article}

\usepackage[numbers]{natbib}

\usepackage[nonatbib,preprint]{neurips_2024}

\usepackage{graphicx}
\usepackage{amsmath}
\usepackage{amssymb}
\usepackage{booktabs}

\usepackage[pagebackref,breaklinks,colorlinks]{hyperref}

\usepackage{caption}
\usepackage{adjustbox}
\usepackage{textcomp}
\usepackage{array} 

\newcommand{\customfootnotetext}[2]{{%
  \renewcommand{\thefootnote}{#1}%
  \footnotetext[0]{#2}}}%

\newcommand{\mailto}[1]{\href{mailto:#1}{\texttt{#1}}}

\usepackage[dvipsnames]{xcolor}

\definecolor{my_green}{RGB}{51,102,0}
\definecolor{my_yellow}{RGB}{255,165,0}
\definecolor{my_red}{RGB}{204, 0, 0}
\definecolor{my_purple}{HTML}{B54DF5}

\newcommand{\red}[1]{\textcolor{red}{#1}}

\newcommand{\green}[1]{\textcolor{my_green}{#1}}

\usepackage{tcolorbox}

\definecolor{myboxcolor}{RGB}{245,245,245} 
\definecolor{myframe}{RGB}{0,0,128} 

\newtcolorbox{mybanner}{
  colback=myboxcolor,
  colframe=myframe,
  boxrule=1pt, % Adjust the border thickness
  left=1pt,
  right=1pt,
  top=1pt,
  bottom=1pt,
}

\newtcolorbox{mybody}{
  colback=myboxcolor,
  colframe=myframe,
  boxrule=1pt, % Adjust the border thickness
  left=1pt,
  right=1pt,
  top=1pt,
  bottom=1pt,
}

\usepackage{wrapfig}
\usepackage{xspace}
\usepackage[utf8]{inputenc} % allow utf-8 input
\usepackage[T1]{fontenc}    % use 8-bit T1 fonts
\usepackage{hyperref}       % hyperlinks
\usepackage{url}            % simple URL typesetting
\usepackage{booktabs}       % professional-quality tables
\usepackage{amsfonts}       % blackboard math symbols
\usepackage{nicefrac}       % compact symbols for 1/2, etc.
\usepackage{microtype}      % microtypography
\usepackage{xcolor}         % colors
\usepackage{subcaption}
\usepackage{pifont}
\usepackage[normalem]{ulem}
\usepackage{amsmath}% http://ctan.org/pkg/amsmath
\usepackage{tablefootnote}
\definecolor{darkgreen}{rgb}{0,0.5,0}

\newcommand{\name}{\textsc{TALC}\xspace} 

\newcommand{\hb}[1]{{#1}}
\newcommand{\ag}[1]{}
\newcommand{\yb}[1]{}
\newcommand{\my}[1]{{#1}}

\title{\name: Time-Aligned Captions for Multi-Scene Text-to-Video Generation}

\author{
 Hritik Bansal$^{1}$
\hspace{0.4em}
 Yonatan Bitton$^{2\textdagger}$\textsuperscript{\textdagger}
\hspace{0.4em}
 Michal Yarom$^{2}$\textsuperscript{\textdagger} \\
\hspace{0.4em}
 \textbf{Idan Szpektor}$^{2*}$ 
\hspace{0.4em}
 \textbf{Aditya Grover}$^{1*}$
\hspace{0.4em}
 \textbf{Kai-Wei Chang}$^{1*}$
\\
$^{1}$University of California Los Angeles \hspace{0.5em} $^{2}$Google Research\\
\url{https://github.com/Hritikbansal/talc}
}

\begin{document}

\customfootnotetext{}{\textsuperscript{\textdagger} Equal Contribution. $^{*}$ Equal Advising. Contact {\mailto{hbansal@ucla.edu},\mailto{yonatanbitton1@gmail.com}}. 
  % $^{1}$UCLA \,
  % $^{2}$Google Research \,
  }

\maketitle

\begin{abstract}

% Recent advances in diffusion-based generative modeling have led to the development of text-to-video (T2V) models that can generate high-quality videos conditioned on a text prompt. 
Most of these text-to-video (T2V) generative models often produce single-scene video clips that depict an entity performing a particular action (e.g., `a red panda climbing a tree'). However, it is pertinent to generate multi-scene videos since they are ubiquitous in the real-world (e.g., `a red panda climbing a tree' followed by `the red panda sleeps on the top of the tree'). To generate multi-scene videos from the pretrained T2V model, we introduce \hb{a simple and effective} \textbf{T}ime-\textbf{Al}igned \textbf{C}aptions (\name) framework. Specifically, we enhance the text-conditioning mechanism in the T2V architecture to recognize the temporal alignment between the video scenes and scene descriptions. For instance, we condition the visual features of the earlier and later scenes of the generated video with the representations of the first scene description (e.g., `a red panda climbing a tree') and second scene description (e.g., `the red panda sleeps on the top of the tree'), respectively. As a result, we show that the T2V model can generate multi-scene videos that adhere to the multi-scene text descriptions and be visually consistent (e.g., entity and background). Further, we finetune the pretrained T2V model with multi-scene video-text data using the \name framework. We show that the \name-finetuned model outperforms the baseline by achieving \hb{a relative gain of $29\%$ in the overall score}, which averages visual consistency and text adherence using human evaluation. 

% The project website is \href{https://talc-mst2v.github.io/}{https://talc-mst2v.github.io/}.
\end{abstract}

\begin{figure*}[h]
\centering
{\includegraphics[width=0.8\textwidth]{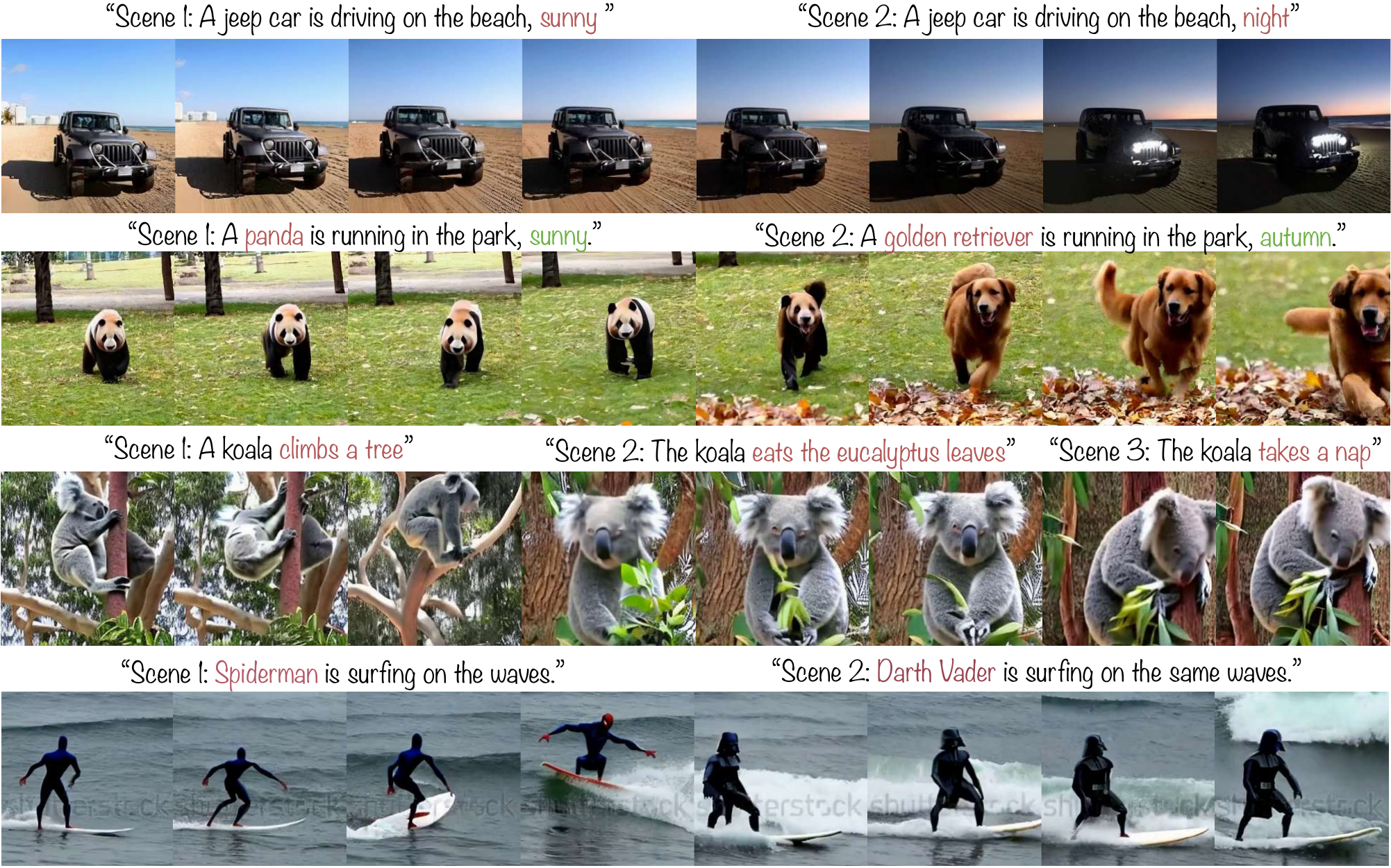}}
\caption{{\bf Examples of multi-scene video generation using the \name framework.} \my{ Our proposed method, \name, enhances scene-level alignment between text and video, facilitating the generation of videos with seamless transitions between textual descriptions while preserving visual consistency. The first two rows are generated using Lumiere \cite{bar2024lumiere} and the last two rows use ModelScope \cite{wang2023modelscope}.}}
\label{fig:talc_examples}
\end{figure*}

\section{Introduction}
\label{sec:introduction}

% \begin{figure}[h]
%  {\includegraphics[width=\linewidth]{images/methods_diagram_v2.pdf}}
% \caption{\small{{\bf Multi-scene video generation methods.} 
% (a) Generating a video by merging descriptions. (b) The resulting video is composed from the merging individual videos from specific scene descriptions. (c) In the \name framework, the generated video is conditioned on the description of scene 1 for the first half of the video frames and on the description of scene 2 for the later video frames. \hb{increase the fontsize of the scene descriptions}}}
% \label{fig:methods_diagram}
% \end{figure}

The ability to generate videos that simulate the physical world has been a long-standing goal of artificial intelligence ~\cite{acharya2018towards,vondrick2016generating,bruce2024genie,sora2024}. In this regard, text-to-video (T2V) models have seen rapid advancements by pretraining on internet-scale datasets of images, videos, and texts \cite{brooks2024video,bar2024lumiere}. Previous work \cite{ho2022imagen,girdhar2023emu,kondratyuk2023videopoet,wang2023modelscope,blattmann2023align,blattmann2023stable} primarily focus on training conditional denoising diffusion probabilistic models \cite{ho2020denoising} on paired video-text data \cite{Bain21,xue2022advancing}. After training, these models allow for video generation by sampling from the trained diffusion model, conditioned on a text prompt. However, most of the open-models such as ModelScope\cite{wang2023modelscope} VideoCrafter \cite{chen2023videocrafter1,chen2024videocrafter2}, OpenSora \cite{opensora} are trained with single-scene video-text dataset \cite{Bain21,wang2023videofactory}, which is widely available and easy to acquire. However, real-world scenarios often require the generation of multi-scene videos from multi-scene descriptions (e.g., \textit{Scene1:} `A koala is napping on a tree.' \textit{Scene2:} `The koala eats leaves on the tree.'). In such cases, the generated video should accurately depict the events in their temporal order (e.g., \textit{Scene2} follows \textit{Scene1}) while maintaining visual consistency, meaning that backgrounds and entities should remain consistent across scenes. While high-performance T2V models such as Sora \cite{sora2024} might be able to generate multi-scene videos, we point out that they are closed-source models trained with massive compute resources and lack sufficient details on the model design, training protocol, and datasets. In this work, we present a complementary approach and tackle the challenge of effectively leveraging the capabilities of base T2V models for multi-scene video generation.

The multi-scene T2V generation differs from long video synthesis where the goal is to either interpolate (few frames to many frames) \cite{girdhar2023emu} or create continuing patterns of the single event in the generated video \cite{blattmann2023align}. Prior work like Phenaki \cite{villegas2022phenaki,kondratyuk2023videopoet} use transformers \cite{vaswani2017attention,anil2023palm} to generate video frames for a given scene autoregressively. However, it is hard for their model to generate multiple scenes reliably as the context length increases with the history of text descriptions and visual tokens \cite{yu2023magvit} of the previous generated videos (e.g., generating \textit{Scene 4} conditioned on the \textit{Scene1, 2, 3} videos and descriptions). Other works \cite{rahman2023make} utilize a latent diffusion model \cite{rombach2022high} to generate video frames autoregressively by conditioning on the entire history of generated videos and scene descriptions. However, the approach is (a) slow due to repeated sampling, (b) generates only one frame per scene description, and (c) shown to work with only limited cartoon characters \cite{li2019storygan,yin2023nuwa} instead of wide range of visual concepts in the real-world. In this work, our goal is to generate multi-scene videos in the end-to-end manner using a diffusion T2V generative model. Prior work \hb{like VideoDirectorGPT} \cite{lin2023videodirectorgpt,long2024videodrafter} generates multi-scene videos by utilizing knowledge of the entity, background, and their movements from large language models \cite{achiam2023gpt}. However, these videos are generated independently for each scene before being merged.

% \input{files/methods_diagram}
% \input{files/video_generation_methods}

% As shown in Figure \ref{fig:methods_diagram}(a), the naive approach to generating a multi-scene video for the scene descriptions would condition the T2V generative model on the merged descriptions. 
% In this setup, the diffusion model processes the entire scene description together, and lacks any information regarding the expected temporal order of events in the generated videos. 
% As a result, we find that this approach leads to poor text-video alignment. 
% As shown in Figure \ref{fig:methods_diagram}(b), an alternative approach generates videos for the individual text descriptions independently and concatenates them in the raw input space along the temporal dimension.
% While this approach achieves good alignment between the scene description and the scene-specific video segment, the resulting video lacks visual consistency in terms of entity and background appearances.

% Moreover, these methods do not offer a way to learn from real-world multi-scene video-text data. 

To remedy these challenges, we propose \name (\textbf{T}ime-\textbf{AL}igned \textbf{C}aptions), a simple and effective framework to generate consistent and faithful multi-scene videos. In particular, our approach conditions the T2V generative model with the knowledge of the temporal alignment between the parts of the multi-scene video and multi-scene descriptions (Figure \ref{fig:architecture}). Specifically, \name conditions the visual representations of earlier video frames on the embeddings of the earlier scene description, and likewise, it conditions the representations of later video frames on the embeddings of the later scene description in the temporal dimension. Additionally, the temporal modules in the T2V diffusion architecture allows information sharing between video frames (the first half and the second half) to maintain visual consistency. Therefore, \name enhances the scene-level text-video alignment while providing the scene descriptions to the diffusion model all at once (Figure \ref{fig:talc_examples}). 

\hb{Prior methods like FreeNoise \cite{qiu2023freenoise} propose a motion injection strategy to address these challenges. However, this method is quite sophisticated and difficult to control due to diverse hyperparameters, such as time-specific motion injection, prompt interpolation, and injection in specific cross-attention layers. In contrast, our approach eliminates these complexities and introduces a straightforward mechanism that significantly boosts performance. Unlike previous work, we also demonstrate that finetuning with \name using real-world multi-scene data can enhance generation capabilities. Specifically, we propose a pipeline to curate multi-scene video data and subsequently finetune the T2V model on this multi-scene data using \name (\S \ref{sec:multi_scene_data_creation}).}

\begin{figure}[h]
% \begin{wrapfigure}{R}{0.55\textwidth}
    \centering
    \includegraphics[width=0.5\linewidth]{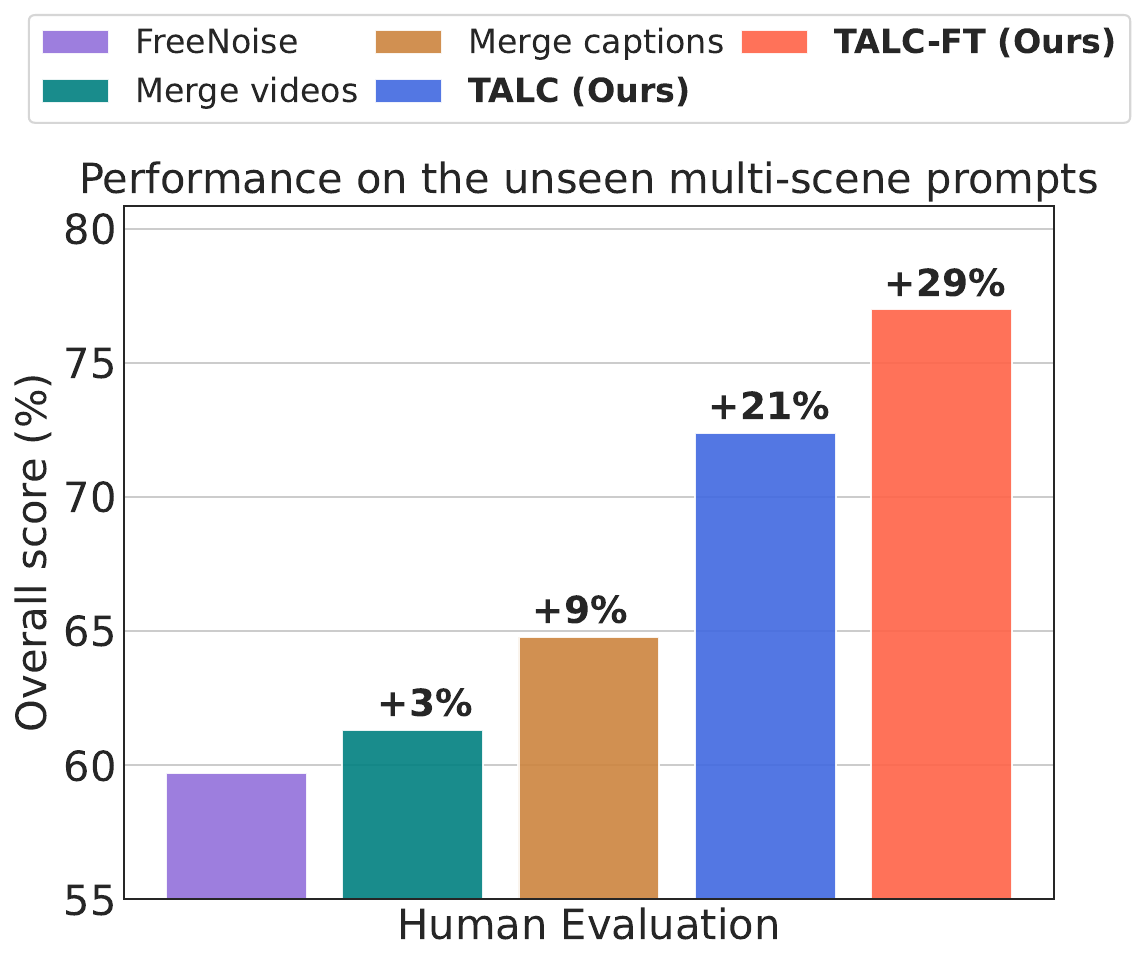}
    \caption{\small{\textbf{Summary of the results.} We compare several baselines and our \name framework with ModelScope \cite{wang2023modelscope} for generating multi-scene videos on the unseen prompts. Specifically, we study the overall score which averages the visual consistency (object and background) and text adherence (video-text alignment). We observe that using \name with the base model i.e., training-free achieves relative gains of $21\%$ in comparison to FreeNoise \cite{qiu2023freenoise}. In addition, we show that multi-scene finetuning with \name (\name-FT) allows larger gains, achieving relative gains upto $29\%$.}}
    \label{fig:pull_figure}
\end{figure}
% \end{wrapfigure}

In our experiments, we assess the visual consistency (background and entity consistency) and multi-scene script adherence of the generated videos from T2V generative models. 
% like Modelscope \cite{wang2023modelscope} and Lumiere \cite{bar2024lumiere}. 
Through our human evaluation, we find that 
% merging scene descriptions leads to high visual consistency but poor text adherence. On the other hand, we observe that merging videos independently achieves the highest text adherence while the visual consistency is compromised. Interestingly, switching to 
\name strikes an effective balance between visual consistency and text adherence, and \hb{outperforms the FreeNoise by achieving relative gains of $21\%$ points on the overall score.} This score represents the average of visual consistency and text adherence scores (Figure \ref{fig:pull_figure}). Furthermore, we construct a multi-scene text-video dataset from real-world videos and fine-tune the T2V generative model using \name. \hb{We show that finetuning with \name outperforms FreeNoise by achieving relative gains of $29\%$ on the overall score (Figure \ref{fig:pull_figure}). Further, we perform automatic evaluation for scalable judgements (\S \ref{sec:automatic_evaluation}) and provide qualitative examples to highlight the benefits of our approach (\S \ref{sec:qualitative_analysis}).} We present the related work in Appendix \S \ref{sec:related_work}.
% \input{files/qualtative_examples}

% On our human evaluation, the generated videos from the \name-finetuned model exhibit higher text adherence than the base model in multi-scene scenarios. Specifically, it outperforms the baseline methods by $15.5$ points on the overall score. 

%In summary, our contributions are:

% \begin{enumerate}
% \item We introduce the \name framework, which enables the generation of multi-scene videos from pretrained Text-to-Video (T2V) diffusion models in an end-to-end manner.
% \item We demonstrate that our framework achieves a good balance between visual consistency and text adherence in the generated videos and outperforms the baseline methods during inference, without requiring any fine-tuning.
% \item Upon fine-tuning the base model with \name on the real-world multi-scene video-text data, we observe an improvement in the overall multi-scene generation capabilities compared to the base model. This establishes the \name framework as an essential inductive bias for multi-scene text-to-video training.
% \end{enumerate}

% \input{files/related_work}
\section{Preliminaries}
\label{sec:background}

% In this work, we focus on generating multi-scene videos from scene descriptions using a diffusion-based Text-to-Video (T2V) generative model. The initial step is to equip the generative model with the knowledge of a wide range of visual concepts and actions. This is achieved during the pretraining stage (\S \ref{sec:diffusion_t2v}). Subsequently, we aim to utilize the base model for multi-scene text-to-video generation task, which we formalize in (\S \ref{sec:mst2vg}). 

% In \S \ref{sec:method}, we propose our \name framework and discuss collection of real-world multi-scene text-video data for finetuning the base T2V model.

\subsection{Diffusion Models for Text-to-Video Generation}
\label{sec:diffusion_t2v}

Diffusion models \cite{ho2020denoising,luo2022understanding} $p_\theta(x)$ are a class of generative models that learn data distribution $p_{data}(x)$. Due to their flexible design, we can train their class-conditional versions to learn class-conditional data distributions $p_{data}(x|y)$ where $y$ is the conditioning variable.
% , that can take various forms such as labels from a dataset or text description accompanying in a video \cite{ramesh2021zero}. 
We assume a dataset $\mathcal{S} \subset \mathcal{V} \times \mathcal{T}$ consisting of pairs of $(V_j, T_j)$ where $V_j \in \mathbb{R}^{L \times 3 \times H \times W}$ is a raw video consisting of $3$ RGB channels, $L$ frames, $H$ height, $W$ width, and $T_j$ is a text caption. We use $\mathcal{V}$ and $\mathcal{T}$ to denote the domain of videos and text, respectively. The aim of T2V generative modeling is to learn the conditional distribution of the videos conditioned on the text $p_{\mathcal{S}}(V_j|T_j)$. In this work, we consider diffusion-based generative models that learn the data distribution via iterative denoising of the input video $z_j \in \mathbb{R}^{L \times C \times H' \times W'}$. Here, $z_j$ can either represent the input video in the raw pixel space $V_j$ \cite{bar2024lumiere} or it can represent the latent representation of the video $z_j = \mathcal{E}(V_j)$ for the latent diffusion models \cite{rombach2022high} where $\mathcal{E}$ is an encoder network \cite{kingma2013auto}.

% These models consists of an encoder module $\mathcal{E}$ and decoder module $\mathcal{D}$ to compress the video $z_j = \mathcal{E}(V_j) \in \mathbb{R}^{L \times C \times H' \times W'}$ and reconstruct the latent $\hat{V}_j = \mathcal{D}(z_j)$, respectively, where $C$ is the number of channels, $H'$ and $W'$ is the height and weight of the latent representation of the video $V_j$. 
% Specifically, the LDM learns to model the data distribution via iterative denoising of the noised latent of the input video. 

Given $z_j$, diffused variable $z_{\tau,j} = \alpha_{\tau}z_j + \beta_\tau\epsilon$ are constructed where $\epsilon \sim \mathcal{N}(0, I)$ where $\alpha_\tau$ and $\beta_\tau$ are sampled from the noise scheduler $p_\tau$ \cite{chen2023importance}. Finally, we train a denoiser network $f_\theta$ \cite{ronneberger2015u,peebles2023scalable} that inputs the diffused variable $z_{\tau}$ and embeddings of the text caption to predict the target vector $y$ where $y$ can be the original noise $\epsilon$, which minimizes the denoising objective \cite{ho2020denoising}: 
\begin{equation}
    \mathbb{E}_{(V_j, T_j) \in S, \tau \sim p_{\tau}, \epsilon \sim N(0, I)}\left[||\epsilon - f_\theta(\tau, z_{\tau,j}, h_j)||_2^2\right]
\end{equation}

where $h_j = \mathcal{H}(T_j)\in \mathbb{R}^{d}$ is the embedding of the text caption $T_j$ where $\mathcal{H}$ is the text embedding model \cite{radford2021learning} and $d$ is the dimension size. 

\subsection{Text Conditioning Mechanism}
\label{sec:text_conditioning_mechanism}

To ensure the effective textual controllability of video generation, the structure of the denoiser networks is equipped with a cross-attention mechanism \cite{wang2023modelscope,girdhar2023emu}. Specifically, it conditions the visual content $z_\tau \in \mathbb{R}^{L \times C \times H' \times W'}$ on the text. To do so, we first \textit{repeat} the text embeddings of the text caption $r_j = R(h_j) \in \mathbb{R}^{L \times d}$ where $R$ is a function that repeats the input text embedding $h_j$ for $L$ times in the temporal dimension. Intuitively, the repeat operation represents that the $L$ frames of the video $z_j$ are semantically aligned with the textual description $T_j$ or its text embedding $r_j$. 

% In \S \ref{sec:method}, we will manipulate this operation to make the model architecture aware of the video-text alignment in the multi-scene scenario. 

These repeated text embeddings $r_j$ are inputs to the spatial attention block as the key and value in the multi-head attention block. The cross-attention enables the intermediate visual features to capture the semantic information that facilitates an alignment between the language and vision embeddings.
% Formally, 
\begin{equation}\label{eq:ori_ca}
    z'_{\tau,j} = CA_{f_{\theta}}(Q=z_{\tau,j}; K = r_j; V = r_j)
\end{equation}
where $CA_{f_{\theta}}$ is the cross attention mechanism with $Q, K, V$ as the query, key, and value, respectively, in the spatial blocks of the denoiser network. Additionally, $z'_{\tau,j}$ is the intermediate representation that is informed with the visual and textual content of the data. In addition to the spatial blocks, the denoiser network also consists temporal blocks that aggregate features across video frames which are useful for maintaining visual consistency in the video. 

\subsection{Multi-Scene Text-to-Video Generation}
\label{sec:mst2vg}

% In many real-world scenarios, such as movies, stories, and instructional videos \cite{zala2023hierarchical}, a video may depict multiple transitions with the same or changing entities, as well as multiple actions or events. In addition, the different video segments often share contextual information such as the background or location. These videos are considered multi-scene videos. 
In this work, we aim to generate multi-scene video $X=\{x_1, x_2, \ldots, x_n\}$ from multi-scene descriptions $Y=\{y_1, y_2, \ldots, y_n\}$ where $n$ are the number of sentences and each sentence $y_j$ is a scene description for scene $j$. Additionally, the index $j$ also defines the temporal order of events in the multi-scene script i.e., we want the events described in the scene $j$ to be depicted earlier than the events described in the scene $k$ where $k>j$. Further, we want the parts of the entire generated video $X$, given by $x_j$, to have high video-text semantic alignment with the corresponding scene description $y_j$. In addition, we expect the appearance of the objects to remain consistent throughout the video unless a change is specified in the description.

% Consider a two-scene description $Y=$ \{`A red panda climbs on a bamboo forest.', `The red panda sleeps peacefully in the treetop.'\}. Here, we need the T2V generative model to synthesize the appearance of the red panda (an entity) that remains consistent throughout the generated video, also referred to as \textit{entity consistency}. In addition, we will expect that the context of the multi-scene video of a forest (a background) to remain consistent, also referred to as \textit{background consistency}. 

\section{\name: Time-Aligned Captions for Multi-Scene T2V Generation}
\label{sec:method}

\subsection{Approach}
\label{sec:main_method}

Most of the existing T2V generative models \cite{wang2023modelscope,chen2023videocrafter1,bar2024lumiere} are trained with large-scale short video-text datasets (10 seconds - 30 seconds) such as WebVid-10M \cite{Bain21}. Here, each instance of the dataset consists of a video and a human-written video description. These videos either lack the depiction of multiple events, or the video descriptions do not cover the broad set of events in the video, instead focusing on the major event shown. As a result, the pretrained T2V generative models only synthesize single video scenes depicting individual events.

We introduce \name, a novel and effective framework to generate multi-scene videos from diffusion T2V generative models based on the scene descriptions. Our approach focuses on the role of text conditioning mechanism that is widely used in the modern T2V generative models (\S \ref{sec:text_conditioning_mechanism}). Specifically, we take inspiration from the fact that the parts of the generated video $x_j$ should depict the events described in the scene description $y_j$. To achieve this, we ensure that the representations for the part of the generated video aggregates language features from the scene description $y_j$. 

\begin{wrapfigure}{R}{0.6\textwidth}
\centering
{\includegraphics[width=0.58\textwidth]{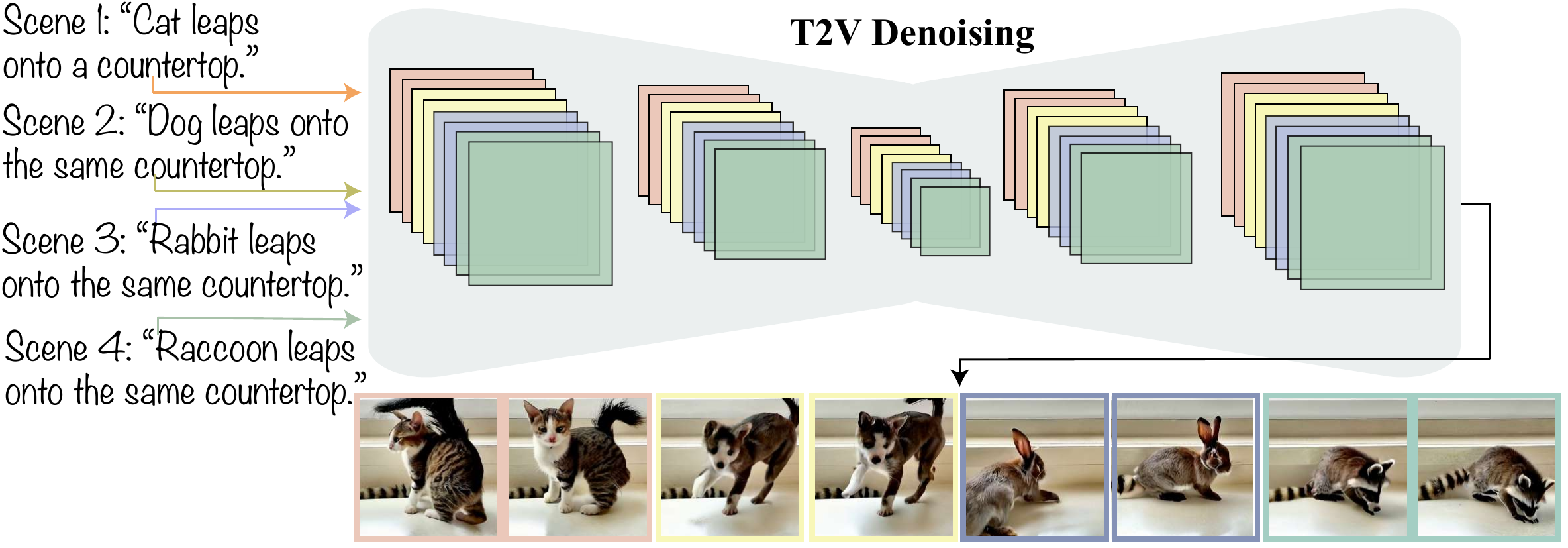}}
\caption{{\bf The architecture of Time-Aligned Captions (TALC).} 
During the generation process of the video, the initial half of the video frames are conditioned on the embeddings of the description of scene 1 ($r_{y_1}$), while the subsequent video frames are conditioned on the embeddings of the description of scene 2 ($r_{y_2}$).}
\label{fig:architecture}
% \end{figure}
\end{wrapfigure}

Consider that we want to generate a multi-scene video $X \in \mathbb{R}^{L \times 3 \times H \times W}$ from the scene descriptions $y_j \in Y$, using a T2V generative model $f_\theta$. Furthermore, we assume that individual video segments $x_j$ are allocated $L/n$ frames within the entire video $X$. Let $z_X = [z_{x_1}; z_{x_2}; \ldots; z_{x_n}] \in \mathbb{R}^{L \times C \times H' \times W'}$ represent the representation for the entire video $X$, and $z_{x_j} \in \mathbb{R}^{(L/n) \times C \times H' \times W'}$ for the $j^{th}$ part of the video that are concatenated in the temporal dimension. In addition, consider $r_{Y} = \{r_{y_1}, \ldots, r_{y_n}\}$ be the set of text embeddings for the multi-scene description $Y$ and $y_j$ be an individual scene description. In the \name framework, the Eq. \ref{eq:ori_ca} is changed to:
\begin{align}
       z'_{\tau, x_j} &= CA_{f_\theta}(Q=z_{\tau, x_j}, K = r_{y_j}, V = r_{y_j}) \\
       z'_{\tau, X} &= [z'_{x_1}; z'_{x_2}; \ldots; z'_{x_n}]
\end{align}
Here, $\tau$ represents the timestamp in the diffusion modeling setup, which is applied during training as well as inference. We illustrate the framework in Figure \ref{fig:architecture}. While \name aims to equip the generative model with the ability to depict all the events in the multi-scene descriptions, the visual consistency is ensured by the temporal modules (attentions and convolution blocks) in the denoiser network. By design, our approach can be applied to the base T2V model without any further training. 

% In \S \ref{sec:experiments}, we show that our approach outperforms relevant baselines (\S \ref{sec:baseline}) for multi-scene video generation in terms of visual consistency and text adherence. 
% \begin{figure}[t]

\subsection{Multi-Scene Video-Text Data Generation}
\label{sec:multi_scene_data_creation}

While our approach generates better multi-scene videos, the text adherence capabilities of the pretrained T2V generative model are limited. This is due to the lack of multi-scene video-text data during its pretraining. 
% Unlike single video-text datasets, the multi-scene video-text datasets are not widely available and are hard to curate for model training. This is attributed to the fact that high-quality caption generation requires human labor which is expensive. 
% Prior work such as ActivityNet \cite{caba2015activitynet} has curated human captions for specific video scenes depicting useful actions in long videos. However, the video scenes are either overlapping or have a large temporal gap between them that will be harmful for natural and smooth variations between the generated multi-scene videos. 
% Hence, the absence of high-quality captions for continuous video scenes in the dataset makes unsuitable for T2V generative training. 
To this end, we create a real-world multi-scene video-text dataset to allow further training of the pretrained T2V models. Specifically, we leverage the capability of the multimodal foundation model, Gemini-Pro-Vision \cite{team2023gemini}, to generate high-quality synthetic data for enhanced video-text training \cite{bansal2023videocon}. Formally, we start with a video-text dataset $\mathcal{M} = \mathcal{A} \times \mathcal{B}$ consisting of pairs of $(A_i, B_i)$ where $A_i$ is a raw video and $B_i$ is the corresponding video description from the dataset. Subsequently, we utilize PySceneDetect library \footnote{\url{https://github.com/Breakthrough/PySceneDetect}} to generate continuous video scenes from $A_i = \{A_{i,1}, A_{i,2}, \ldots, A_{i,m}\}$ where $m$ is the number of scene cuts in the video. 
% A similar approach was used in a prior work \cite{blattmann2023stable} to detect scene changes in the video data. 
Further, we sample the middle video frame $F_{i,j}$ as a representative of the semantic content in the video scene $A_{i,j}$. Finally, we input all the video frames $F_i = \{F_{i,1}, \ldots, F_{i,m}\}$ for a single video $A_i$ and the entire video caption $B_i$ to Gemini-Pro-Vision. Specifically, the model is prompted to generate high-quality captions for each of the frames $F_{i,j}$ such they form a coherent narrative guided by the common caption $B_i$. We provide the prompt provided to the multimodal model in Appendix \S \ref{app:prompt_gemini_pro_vision}. In Figure \ref{fig:gemini_multi_image_data} we provide an instance for the multi-scene video-text data generation. We highlight that higher-quality multi-scene datasets would enhance the performance of the models with \name framework. We provide the more details in Appendix \ref{app:video_data_sources}. \footnote{We note that the parts of this pipeline are also utilized in a concurrent work, ShareGPTVideo \cite{chen2024sharegpt4video}. However, their approach is focused on long-form video generation pretraining, while we are focus on multi-scene video generation for finetuning. }

% In \S \ref{sec:multi_scene_data_creation}, we discuss the construction of a real-world multi-scene video-text dataset. Finally, we finetune the base T2V model on the collected data and show that the finetuned model, combined with the \name framework, enhances the text adherence capabilities of the base model for the multi-scene video generation task.

% \hb{need a figure -- Yonatan}\yb{yes it is there now and I see you refer to it}

\section{Evaluation}
\label{sec:evaluation}

% In this section, we describe the evaluation scheme for videos generated from multi-scene text descriptions. 

% First, we describe the evaluation metrics that we aim to assess in this work (\S \ref{sec:eval_metric}). Then, we generate multi-scene descriptions for a diverse set of tasks (\S \ref{sec:eval_task}). Finally, we present the details for automatic and human evaluation of the generated videos (\S \ref{sec:eval_evaluator}).

\subsection{Metrics}
\label{sec:eval_metric}

The ability to assess the quality of the generated multi-scene videos is a challenging task itself. As humans, we can judge the multi-scene videos across diverse perceptual dimensions \cite{huang2023vbench} that the existing automatic methods often fails to capture \cite{bugliarello2024storybench}. Following \cite{lin2023videodirectorgpt}, we focus on the visual consistency of the generated video, and text adherence capabilities of the T2V models. 

% Here, we present the metrics with the aspects that they intend to assess in the generated video for multi-scene text description.

\textbf{Visual Consistency.} This metric aims to assess the (entity or background) consistency between the frames of the multi-scene videos. Here, the \textbf{entity consistency} aims to test whether the entities in the multi-scene video are consistent across the video frames. For instance, the appearance of an animal should not change without a change described in the text description. In addition, the \textbf{background consistency} aims to test whether the background of the multi-scene video remains consistent across the video frames. For instance, the room should not change without a change in text description.

\textbf{Text Adherence.} This metric aims to test whether the generated video adheres to the multi-scene text description. For instance, the events and actions described in the text script should be presented in the video accurately, and in the correct temporal order. 

In our experiments, we compute the visual consistency and text adherence with the human and automatic evaluators. Further, we compute the overall score, which is the average of the visual consistency and text adherence scores. In addition, we also assess the visual quality of the generated videos using human evaluation to understand whether the video contains any flimsy frames, shaky images, or undesirable artifacts (Appendix \ref{app:visual_quality}). 

\subsection{Task Prompts}
\label{sec:eval_task}

Here, we curate task prompts for diverse scenarios to holistically assess the quality of the videos.

\textbf{Single character in multiple visual contexts (S1).} In this scenario, we instruct an LLM, GPT-4 \cite{achiam2023gpt}, to create a coherent script consisting of four scenes. Each scene features a specific animal character performing diverse activities in every scene. This task assesses the capability of the T2V model to generate consistent appearance of the entity and its background while adhering to the different actions (or events) described in the multi-scene text script. 
% For instance, a generated script could be `Scene 1: A red panda is climbing a tree. Scene 2: The red panda eats the leaves on the tree. Scene 3: The red panda lies down on the branch of the tree. Scene 4: The red panda sleeps on the branch'. In total, we generate $100$ prompts in this scenario.

\textbf{Different characters in a specific visual context (S2).} In this scenario, we instruct a language model, GPT-4, to create a coherent script consisting of four scenes. Each scene features different animal characters engaging in the same activity in every scene \cite{villegas2022phenaki}. This task assesses the capability of the T2V model to generate consistent appearance of the background while adhering to the appearance of the different characters in the multi-scene text script. 

% For instance, a generated script could be `Scene 1: A cat leaps onto countertop. Scene 2: A dog leaps onto the same countertop. Scene 3: A rabbit leaps onto the same countertop. Scene 4: A raccoon leaps onto the same countertop'. In total, we generate $100$ prompts in this scenario.

\textbf{Multi-scene captions from real videos (S3).} Here, we aim to assess the ability of the model to generate multi-scene videos for open-ended prompts that are derived from real-world videos. This task also assesses the ability of the T2V model to generate consistent appearances of the entity and its background while adhering to multi-scene descriptions. Specifically, we use our multi-scene video-text data generation pipeline (\S \ref{sec:multi_scene_data_creation}) to create such prompts for the real videos from the test splits of the video-text datasets. In total, we generate $100$ prompts in this scenario. We present a sample of the various task prompts in the Appendix \S \ref{app:task_prompts}. We present the details about the human and automatic evaluators in \S \ref{sec:eval_evaluator}.

% For example, a multi-scene text script could be `Scene 1: A beauty vlogger introduces her skincare routine. Scene 2: She applies a serum to her face, smoothing it in'. 

\subsection{Evaluation Setup}
\label{sec:eval_setup}

% Since merging captions (\S \ref{sec:baseline}) and \name (\S \ref{sec:main_method}) methods input the entire multi-scene text description at once, the quality of the video generated by these methods is influenced by the number of scenes described in the text script. Hence, 
We compute the performance of the baselines and \name by averaging the scores assigned to videos generated for two, three, and four scenes. Additionally, we report on visual consistency by averaging the performance across the entity and background consistency metrics. Here, the entity consistency scores are calculated for the task prompts S1 and S3 (since S2 aims to change the characters across scenes), and the background consistency and text adherence scores are computed for all the task prompts. We also evaluate the impact of \name-based finetuning for single-scenes in Appendix \S \ref{sec:single_scene_eval}. 
\begin{table*}[t]
\centering
\caption{\small{\textbf{Human evaluation results.} We present the human evaluation results for the overall score for several baselines (e.g., FreeNoise, merge videos, and merge captions) and \name framework for the ModelScope generative model. Specifically, we find that the \name with the base model outperforms the FreeNoise by achieving the relative gain of $21\%$ on the overall score. In addition, finetuning ModelScope with \name framework enables better multi-scene video generation by achieving the highest overall score. Overall, \name strikes a good balance between visual consistency and text adherence.}}
\label{tab:human_eval}
\begin{tabular}{l|c|cc}
\hline
                    & \textbf{Overall score} & \textbf{Visual consistency} &\textbf{ Text adherence} \\\hline
FreeNoise \cite{qiu2023freenoise}           & 59.7          & 77.0               & 42.5           \\
Merge videos        & 61.3 (\green{$+2.6\%$})          & 55.0               & \textbf{67.5}           \\
Merge captions      & 64.8 (\green{$+8.5\%$})         & \textbf{96.5}               & 33.0             \\
Finetuning w/ Merge captions & 61.1  (\green{$+2.3\%$})        & 80.0               & 42.3           \\\hline
\name                & 72.4 (\green{$+21\%$})          & 92.3               & 52.5           \\
Finetuning w/ \name           & \textbf{76.8} (\green{$+29\%$})          & 86.4               & 67.2        \\\hline
\end{tabular}
\end{table*}

\section{Experiments}
\label{sec:experiments}

\subsection{Text-to-Video Generative Models}
\label{sec:setup}

% Here, we describe the T2V models (\S\ref{sec:t2v_model}) tested in this work and the real-world text-video dataset utilized to create multi-scene video-text data (\S\ref{sec:rw_t2v}).

% \subsection{Text-to-Video Generative Models}
% \label{sec:t2v_model}

% In this work, we utilize ModelScope \cite{wang2023modelscope} and Lumiere \cite{bar2024lumiere} T2V models for multi-scene video generation. 
% Here, ModelScope is an open-source T2V model with 1.7 billion parameters including the video encoder, text encoder, and denoising U-net network. 
% Specifically, it is trained to generate 16 video frames on the mix of WebVid \cite{Bain21} video-text dataset and LAION \cite{laion} image-text dataset. 
We perform most of our experiments on ModelScope \cite{wang2023modelscope} due to its easy-of-access and adoption in prior works \cite{lin2023videodirectorgpt}. In addition, we also include Lumiere-T2V, a model that leverages space-time U-Net denoising networks to generate high-quality videos. In this work, we include early experiments with Lumiere to showcase the flexibility of the \name approach on diverse models. \hb{We perform human evaluation for ModelScope, and automatic evaluation for both ModelScope and Lumiere.}

\textbf{Base model with \name.} As described in \S \ref{sec:main_method}, our approach modifies the traditional text-conditioning mechanism to be aware of the alignment between text descriptions and individual video scenes. By design, the \name framework can be applied to the base T2V model during inference, without any multi-scene finetuning. Here, we generate $16$ frames per scene from ModelScope and $80$ frames per scene from Lumiere. We provide more details in Appendix \S \ref{sec:inference_details}.

% Thus, we compare the performance of the multi-scene videos generated from ModelScope and Lumiere T2V base models under three settings: merging captions, merging videos, and \name. In this setting, 

\textbf{Finetuning with \name.} Since the base model is pretrained with single-scene data, we aim to show the usefulness of \name framework when we have access to the multi-scene video-text data. To this end, we finetune ModelScope on the multi-scene video-text data (\S \ref{sec:multi_scene_data_creation}) with \name framework. As a pertinent baseline, we also finetune the ModelScope without \name framework by naively merging the scene-specific captions in the raw text space. In this setting, we finetune the T2V model with $8$ frames per scene and the maximum number of scenes in an instance is set to $4$. We provide further details on the finetuning setup in Appendix \S \ref{sec:finetuning_details}. The inference settings are identical to the prior method of generating videos from the base model without finetuning.

In this section, we present the results for the baselines and \name framework averaged over a diverse task prompts and multiple scenes using automatic evaluation (\S \ref{sec:automatic_evaluation}) and human evaluation (\S \ref{sec:human_evaluation}). Finally, we provide qualitative examples for the multi-scene generated videos to showcase the usefulness of our approach (\S\ref{sec:qualitative_analysis}).

\subsection{Baselines}
\label{sec:baseline}

% Here, we describe the baseline methods that could be used to generate videos for the multi-scene descriptions from a given diffusion text-to-video generative model.

\paragraph{Merge Captions} 
% \label{sec:merge_captions}

In this setup, we create a single caption by merging all the multi-scene descriptions. 
% Specifically, the multi-scene descriptions $Y = \{y_1, y_2, \ldots, y_n\}$ can be written as a single prompt `$\mathcal{P} = y_1. \text{Then, } y_2. \ldots \text{Then, } y_n.$' 
% For instance, the two-scene description $Y=$ \{`A red panda climbs on a bamboo forest.', `The red panda sleeps peacefully in the treetop.'\} will change to $\mathcal{P} = $ `A red panda climbs on a bamboo forest. Then, the red panda sleeps peacefully in the treetop.' 
% Subsequently, we generate a video from the T2V model $f_\theta$ by conditioning it on $\mathcal{P}$.
While this approach mentions the temporal sequence of the events in a single prompt, the T2V model does not understand the temporal boundaries between the two events. Specifically, 
% the Eq. \ref{eq:ori_ca} suggests that 
the visual features for all the frames will aggregate information from the entire multi-scene description, at once, without any knowledge about the alignment between the scene description and its expected appearance. 

% In our experiments (\S \ref{sec:experiments}), we find that this baseline achieves high visual consistency (e.g., the `red panda' entity and `forest' background would appear similar) but poor text adherence (e.g., only depict `red panda climbing' but not `red panda sleeping').

\paragraph{Merge Videos}
% \label{sec:merge_videos}

In this setup, we generate videos for each scene description individually and merge them in the raw input space. 
% Formally, the individual scene description $y_i$ conditions the T2V model $f_\theta$ to generate the parts of the multi-video $x_i$. Finally, we stitch the individual videos together to synthesize the entire video $X = {x_1, x_2, \ldots, x_n}$. 
In this process, the parts of the multi-scene video closely adhere to the scene descriptions, leading to high text fidelity. However, since the generated videos do not have access to all the multi-scene descriptions (e.g., the video for Scene 2 is not informed about Scene 1), the visual consistency across the entire video is quite poor.

\hb{\paragraph{FreeNoise} We also consider a more sophisticated approach, FreeNoise \cite{qiu2023freenoise}. Specifically, it reschedules a sequence of noises for long-range correlation and perform temporal attention over them by window-based fusion. In addition, it includes motion injection method to support the
generation of videos conditioned on multiple text prompts. Due to its complexity, this method involves setting various hyperparameters, including the denoising step at which motion injection should be activated, the cross-attention layers where prompts are injected, and the frames between which the prompt representations should be interpolated. In this work, we evaluate the publicly available implementation with VideoCrafter \cite{chen2023videocrafter1} on our unseen prompts.\footnote{https://github.com/AILab-CVC/FreeNoise}} \hb{While Gen-L-Video \cite{wang2023gen} has been shown to perform worse than FreeNoise, we conduct a qualitative analysis to emphasize the robustness of our method against it. (\S \ref{sec:qualitative_analysis}).\footnote{Other approaches such as VideoDirectorGPT \cite{lin2023videodirectorgpt} and Phenaki \cite{villegas2022phenaki} are not publicly available.}}

\subsection{Human Evaluation}
\label{sec:human_evaluation}

\paragraph{\name achieves the best performance in human evaluation.} We compare the performance of the baselines and \name framework using human evaluation in Table \ref{tab:human_eval}. \hb{We find that \name applied to ModelScope (MS) outperforms FreeNoise by achieving a relative gain of $21\%$ on the overall score. In addition, finetuning MS with \name framework increases the video generation capability, leading to relative gain of $27\%$ over FreeNoise in the overall score. The poor performance of FreeNoise can be attributed to the inability to introduce new content and limited motions as the multiple scenes ($>2$) are requested. We provide some qualitative examples for FreeNoise to highlight its limitations in Appendix \ref{app:freenoise_failure}.} 
In addition, we find that using \name framework in the base model outperforms the merging captions and merging video methods with the base model by $7.6$ points and $11.1$ points, respectively, on the overall score. Further, we note that \name-finetuned model outperforms the merging captions and merging video methods with the base model by $12$ points and $15.5$ points, respectively, on the overall score. 

Additionally, we observe that the merging captions with the base model achieves the highest visual consistency score of $96.5$ points while it is the lowest for merging videos generated from the base model. Our results indicate that merging videos independently for the individual scene descriptions does not preserve the background and entity appearances across the different frames. Further, we note that the text adherence of the \name-finetuned and \name-base model is better than merging captions-finetuned and merging captions-base model, respectively. \hb{The high text adherence for merging videos can be attributed to its design where individual video scenes adhere to the scene-specific descriptions well. Overall, our empirical findings highlight that our simple framework can enable robust multi-scene video generation with finetuning. We present the results for the visual quality metric in Appendix \ref{app:visual_quality}.}

\begin{table}[h]
\centering
\caption{\small{\textbf{Automatic evaluation results for ModelScope.} We compare the performance of the baselines and \name framework using the automatic evaluation for ModelScope generative model. Similar to the human evaluation, we observe that our simple approach achieves the best overall score with the base model (\name) and finetuned model (Finetuning w/ \name). We abbreviate visual consistency as VC, and text adherence as TA.}}
\label{tab:auto_modelscope}
% \resizebox{\linewidth}{!}{%
\begin{tabular}{l|c|cc}
\hline
                    & \textbf{Overall score} & \textbf{VC} &\textbf{TA} \\\hline
Merge captions      & 61.7          & \textbf{91.0}               & 32.4           \\
Merge videos        & 67.5  (\green{$+9.4\%$})        & 65.0               & \textbf{70.0}             \\
Finetuning w/ Merge captions               & 57.3 (\red{$-7.1\%$})           & 77.0               & 37.5           \\\hline
\name & 68.6 (\green{+11.2\%})          & 89.9               & 47.2           \\
Finetuning w/ \name & \textbf{75.6}  (\green{+22.5\%})        & 89.0               & 62.3        \\\hline
\end{tabular}
%}
\end{table}
\begin{table}[h]
\centering
\caption{\small{\textbf{Automatic evaluation results for Lumiere.} We present the results for the comparison between the baselines and \name for the Lumiere video generative model. Specifically, our results indicate that \name achieves the highest overall score, by achieving $7.1\%$ relative gains over merge captions on the overall score. This indicates that \name is a flexible strategy that can be applied to diverse video generative models. We abbreviate visual consistency as VC, and text adherence as TA.}}
\label{tab:auto_lumiere}
% \resizebox{\linewidth}{!}{%
\begin{tabular}{l|c|cc}
\hline
               & \textbf{Overall score} & \textbf{VC} & \textbf{TA} \\\hline
Merge captions & 64.4          & 94.7               & 34             \\
Merge videos   & 66.5 (\green{+3.2\%})         & 68.0               & \textbf{65.0}             \\\hline
\name           & \textbf{69.0} (\green{+7.1\%})   & \textbf{97.8}               & 40.0    \\\hline
\end{tabular}
%}
\end{table}

\subsection{Automatic Evaluation}
\label{sec:automatic_evaluation}

We compare the performance of the baselines with the \name framework for ModelScope and Lumiere using the automatic evaluation in Table \ref{tab:auto_modelscope} and \ref{tab:auto_lumiere}, respectively.

\paragraph{\name outperforms the baselines without any finetuning.} In Table \ref{tab:auto_modelscope}, we find that the overall score, average of visual consistency and text adherence, of the multi-scene videos generated using the base ModelScope with \name ($68.6$ points), outperforms the overall score achieved by the videos generated using merging captions ($61.7$ points) and merging videos ($67.5$ points) with the base ModelScope. In addition, we observe that the text adherence using \name outperforms merging captions by $14.8$ points, while the text adherence is the highest with a score of $70$ points using merging videos. 
% This can be attributed to the design of the merging videos baseline where individual video scenes adhere to the scene-specific descriptions well. 
% % Hence, merging videos independently approach can be viewed as an upper bound on the text adherence metric.
In Table \ref{tab:auto_lumiere}, we observe similar trends for the Lumiere T2V generative model. Specifically, we find that the overall score for \name outperforms merging captions and merging videos by $4$ points and $2$ points, respectively. In addition, we observe that merging captions and \name achieve a high visual consistency score while merging videos independently has poor visual consistency. Further, we find that \name outperforms merging captions by $5$ points on text adherence, while merging videos achieves the highest text adherence $65$ points. This highlights that the model more easily generates multi-scene videos that adhere to individual text scripts, whereas adherence to the text diminishes when the model is given descriptions of multiple scenes all at once.

\paragraph{Finetuning with \name achieves the best performance.} 
% Earlier, we evaluated the usefulness of the \name framework with the base model. However, the base models are trained with the single-scene video-text data that might limit their capability for multi-scene video generation. To alleviate this issue, we finetune ModelScope T2V model on the multi-scene video-text data (\S \ref{sec:multi_scene_data_creation}). Specifically, we finetune the model using the merging captions method and \name framework, independently. 

In Table \ref{tab:auto_modelscope}, we find that finetuning with \name achieves the highest overall score of $75.6$ points in comparison to all the baselines. Specifically, we observe that the visual consistency does not change much with finetuning using the \name method ($89.9$ points vs $89$ points). Interestingly, we observe that finetuning with merging captions reduces the visual consistency by a large margin of $14$ points. This can be attributed to the lack of knowledge about the natural alignment between video scenes and individual scene descriptions, which gets lost during the merging of captions. Additionally, we find that the text adherence of the \name-finetuned model is $15.1$ points more than the text adherence of the \name-base model. This highlights that finetuning a T2V model with multi-scene data helps the most with its text adherence capability.

\vspace{-0.3cm}

\paragraph{Fine-grained Results.} To perform fine-grained analysis of the performance, we assess the visual consistency and text adherence scores for the baselines and \name framework across diverse task prompts and number of scenes on ModelScope. We present their results in Appendix \S \ref{sec:fine_grained_analysis}. In our analysis, we find that finetuning with \name achieves the highest overall score over the baselines across all the scenarios. In addition, we notice that the highest performance is achieved in the scenario that consist of the different entities in a specific visual context. Further, we observe that the performance of the all the methods reduces when the task prompts get more complex i.e., multi-scene captions from real videos. In addition, we observe that finetuning with \name achieves the highest overall score over the baselines across all the number of scenes. Specifically, we observe that the performance of the merging captions and \name framework reduces as the number of scenes being generated increases. Overall, we show that the \name strikes a good balance between visual consistency and text adherence to generate high-quality multi-scene videos. 

\vspace{-0.3cm}
\subsection{Qualitative Analysis}
\label{sec:qualitative_analysis}

\my{ We provide qualitative examples of generating multi-scene videos using \name, FreeNoise and Gen-L-Video in Appendix Figure \ref{fig:time_aligned_examples} and Figure \ref{fig:modelscope_ex}. Our analysis demonstrates that all methods are capable of generating multi-scene videos that exhibit a high degree of text adherence. However, the primary distinction lies in the quality of the videos.} \my{FreeNoise is capable of producing videos characterized by superior quality and visual coherence. Nonetheless, the motion within these videos is notably limited, resulting in a relatively static scene.} \my{Gen-L-Video can generate videos that incorporate motion. However, visual consistency is not maintained throughout the video. For instance, in the provided example depicting a man engaged in both surfing and skiing, the man's appearance undergoes noticeable changes across the video.} \my{Overall, we observe that \name is capable of generating realistic multi-scene videos that not only exhibit high textual adherence but also maintain visual consistency. Furthermore, the transitions between scenes are smooth and natural.}
\vspace{-0.3cm}
\section{Conclusion}

We introduced \name, a simple and effective method for improving the text-to-video (T2V) models for multi-scene generation. Specifically, it incorporates the knowledge of the natural alignment between the video segments and the scene-specific descriptions. Further, we show that \name-finetuned T2V model achieve high visual consistency and text adherence while the baselines suffer from one or both of the metrics. Given its design, our framework can be easily adapted into any diffusion-based T2V model. An important future direction will be to scale the amount of multi-scene video-text data and deploy \name framework during pretraining of the T2V models.

\section{Acknowledgement}

We would like to thank Ashima Suvarna for providing feedback on the draft. Hritik Bansal is supported in part by AFOSR MURI grant FA9550-22-1-0380.

\bibliographystyle{plain}
\bibliography{egbib}

\clearpage

\appendix

\section{Related Work}
\label{sec:related_work}
% \red{rephrased section}
\paragraph{Text-to-Video Generative Modeling:}
% Text-to-video (T2V) synthesis has evolved from early GAN-based models like VGAN~\cite{vondrick2016generating} and MoCoGAN~\cite{tulyakov2018mocogan}, which generated short, single-scene videos, to transformer-based architectures such as CogVideo~\cite{hong2022cogvideo} and VideoGPT~\cite{yan2021videogpt}, which enhance video content complexity but remain limited to single scenes. 

Diffusion models like Imagen Video~\cite{ho2022imagenvideo} represent a significant advancement in T2V synthesis, yet generating multi-scene videos that realistically capture the complexity of the physical world~\cite{acharya2018towards,vondrick2016generating,bruce2024genie} remains challenging. Recent research has attempted longer video generation, but limitations persist. Phenaki~\cite{villegas2022phenaki} targets arbitrary-length videos with a focus on temporal coherence, though its evaluation is constrained by the lack of publicly available code. VideoDirectorGPT~\cite{lin2023videodirectorgpt}, DirecT2V~\cite{hong2023direct2v}, and Free-bloom~\cite{huang2024free} employ zero-shot approaches for multi-scene generation, contrasting with our fine-tuned method for enhanced performance. Gen-L-Video~\cite{wang2023gen} uses iterative denoising to create consistent videos by aggregating overlapping short clips, but it typically generates videos with a maximum of two scenes, whereas our method supports up to four distinct scenes. StreamingT2V~\cite{henschel2024streamingt2v} employs an auto-regressive, streaming-based approach for continuous video generation. While these methods contribute to long video generation, they often struggle with maintaining visual consistency and adherence to multi-scene textual descriptions, which are critical for storytelling.

% \paragraph{Image-to-Video Animation:}
% Approaches like Lumiere~\cite{bar2024lumiere} and Make-a-Video~\cite{singer2022make} generate multi-scene videos by first creating images from text and then animating them. However, these methods often lack narrative consistency across scenes, resulting in disjoint storylines. Works like Emu Video~\cite{girdhar2023emu} and Dynamicrafter~\cite{xing2023dynamicrafter} highlight the difficulty of achieving smooth transitions and coherent storytelling across multiple scenes using this technique.

\paragraph{Multi-Scene Video Generation:}
Efforts like Phenaki~\cite{villegas2022phenaki} and Stable Video Diffusion~\cite{blattmann2023stable} push the boundaries of text-driven generation by scaling latent diffusion models. Dreamix~\cite{molad2023dreamix} and Pix2Video~\cite{ceylan2023pix2video} utilize diffusion models for video editing and animation, while methods like MCVD~\cite{voleti2022mcvd} and VDT~\cite{lu2023vdt} focus on improving temporal consistency in longer videos. Despite these advances, generating multi-scene videos that accurately reflect complex narratives with high visual fidelity remains difficult, as shown by ongoing research in projects like VideoPoet~\cite{kondratyuk2023videopoet}, ModelScope~\cite{wang2023modelscope}, and Make-A-Scene~\cite{gafni2022make}. \name addresses the challenges of multi-scene video generation by enhancing visual consistency and text adherence across scenes. Unlike zero-shot methods, \name leverages fine-tuning on a curated multi-scene video dataset. We also propose a comprehensive evaluation protocol, combining human evaluation and automated assessments using GPT-4V, to ensure robust narrative coherence in the generated videos.

\section{Evaluator}
\label{sec:eval_evaluator}

% In this work, we devise an automatic evaluation framework and perform human evaluation to assess the quality of the multi-scene generated videos.

\textbf{Human Evaluation.} Here, we use the annotators from Amazon Mechanical Turk (AMT) to provide their judgements for the generated videos. Specifically, we choose the annotators that pass a preliminary qualification exam. Subsequently, they assess the multi-scene generated videos along the dimensions of entity and background consistency, text adherence, and visual quality. For each metric, the multimodal model assigns one of three possible response $\{yes = 1, partial = 0.5, no = 0\}$. For instance, $yes$ for the entity consistency metric implies that the video frames sampled from the generated video have consistent appearance of the entity described in the multi-scene script. We present the screenshot of the UI in Appendix \S \ref{sec:human_ss}.

\textbf{Automatic Evaluation.} Here, we utilize the capability of a large multimodal model, GPT-4-Vision \cite{gpt4v}, to reason over multiple image sequences. First, we sample four video frames, uniformly, from each scene in the generated video (e.g., 8 videos frames for two-scene video). Then, we prompt the multimodal model with the temporal sequence of video frames from different scenes and the multi-scene text description. The model is instructed to judge the generated videos for visual consistency and text adherence, similar to the human evaluation. In this work, we do not utilize any existing video-text alignment models \cite{xu2021videoclip,bansal2023videocon} for evaluating text adherence as they are trained on single-scene video-text datasets. \hb{We find that the agreement between the automatic evaluation and human evaluation is $77\%$}. We present the automatic evaluation prompt in Appendix \S \ref{app:automatic_eval_prompt}.

\begin{figure*}[h]
    \centering
    \includegraphics[width=0.8\linewidth]{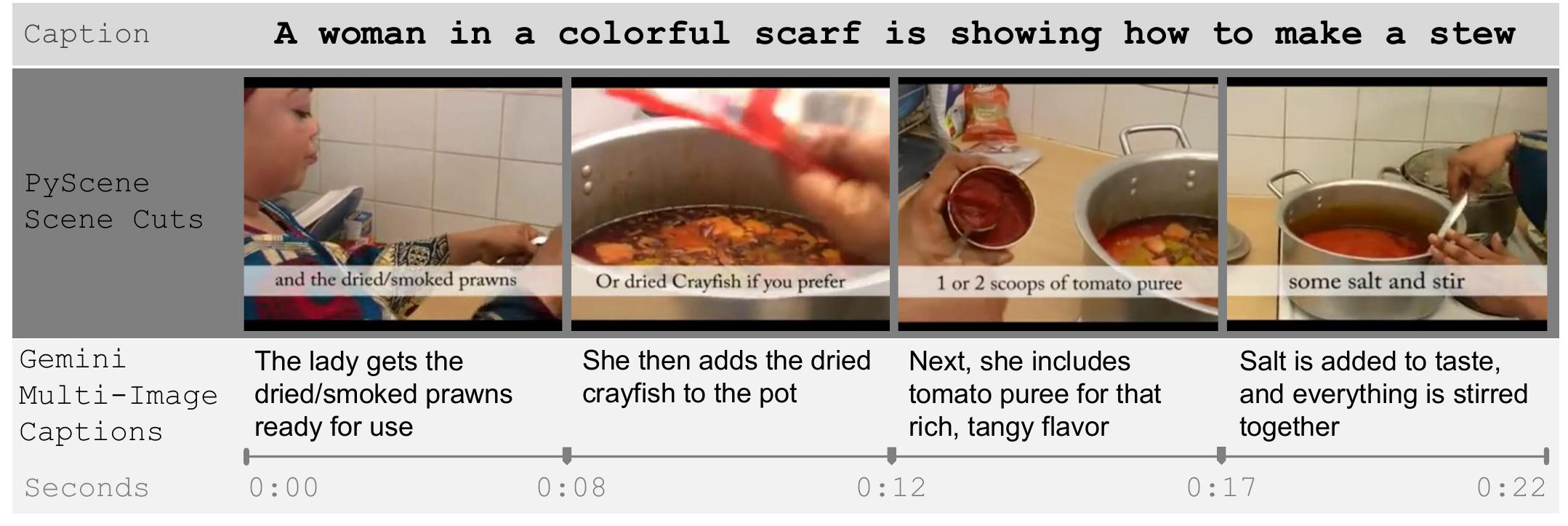}
    \caption{\textbf{Our approach for generating time-aligned video captions.} The process begins with PyScene cuts identifying the boundaries of distinct scenes within a video. Keyframes are then selected from the median of each scene. These frames are processed collectively through the Gemini model to produce multi-image captions that maintain narrative continuity by contextualizing each scene within the video's overall sequence.}
    \label{fig:gemini_multi_image_data}
\end{figure*}

\section{Multi-scene video data sources}
\label{app:video_data_sources}

To construct a multi-scene video-text dataset, we utilize existing dataset that include natural (real) videos and associated high-quality human-written captions that summarize the entire video. Specifically, we choose MSR-VTT \cite{xu2016msr} and VaTeX \cite{wang2019vatex}. Most of the videos in MSR-VTT are 10-30 seconds long while VaTeX consists 10 seconds long videos. In addition, each video in MSR-VTT and VaTeX consists 20 captions and 10 captions, respectively, out of which one is selected at random for multi-scene data generation. As described above, a single video is cut into multiple video segments using Pyscene library. In our experiments, we retain the first four video segments and discard any additional segments if the library generates more than four. Since the accuracy of the multi-scene captioning and the computational demands during finetuning are influenced by the number of scenes, we opt to limit the scene count to four for our experiments. However, future work could employ similar methodologies to scale the number of scenes, given more computing power and advanced multi-scene captioning models. We provide the data statistics for the final multi-scene data in Appendix \S \ref{sec:multi_scene_data_stats}.

\section{Visual quality of the generated videos.}
\label{app:visual_quality}

\begin{table}[h]
\centering
\caption{\textbf{Human evaluation results on the visual quality of the generated videos from ModelScope.} We observe that the visual quality of the generated videos are close to each other for the base model. However, finetuning the model with merging captions reduces the video quality by a large margin while \name-finetuned model retains the video quality.}
\label{sec:video_quality_human}
% \resizebox{\linewidth}{!}{%
\begin{tabular}{cc}
\hline
\textbf{Method}                    & \textbf{Quality} \\
\hline
FreeNoise & 80 \\
Merge captions & 80.5          \\
Merge videos           & 86.5\\
Finetuning w/ Merge captions & 63.4          \\\hline
\name                   & 84.5          \\
Finetuning w/ \name          & 83.3       \\\hline  
\end{tabular}%
% }
\end{table}

We compare the visual quality of the generated videos using human evaluation in Table \ref{sec:video_quality_human}. We find that the visual quality of videos generated from the base model ranges from $80-86.5$ using the baselines and \name framework. However, we observe that the visual quality of generated videos is quite poor for the model finetuned with merging captions with a score of $63.4$ points. This highlights that finetuning a T2V model with multi-scene video-text data by naively merging the scene-specific descriptions in the raw text space leads to undesirable artifacts in the generated video. Finally, we find that the \name-finetuned model ($83.3$) achieves a video quality score similar to that of the \name-base model ($84.5$), indicating that our finetuning data preserves the visual quality observed during the model's pretraining.

\section{Prompt for Multi-Scene Caption Generation}
\label{app:prompt_gemini_pro_vision}

We present the prompt used to generate multi-scene captions using large multimodal model, Gemini-Pro-Vision, in Figure \ref{tab:caption_generation}.  In particular, we utilize Gemini-Pro-Vision since it can reason over multiple image sequences. Specifically, we provide the multimodal model with a video frame from each of the segmented videos and the single caption for the entire video present in the original video-text datasets.

\begin{figure*}[h]
\centering
\resizebox{\linewidth}{!}{

\begin{tabular}{p{1.3\linewidth}}

\toprule
Your task is to create captions for a series of images, each taken from different video scenes. For every image shown, craft a 7-10 word caption (7-10 words) that precisely describes what's visible, while also linking these captions into a fluid, engaging story. A common caption will be given to help guide your narrative, ensuring a smooth transition between scenes for a cohesive story flow. Remember to not hallucinate in your responses.\\\\
    
Common Caption: \{caption\}\\
\bottomrule
\end{tabular} }

\caption{Prompt to generate multi-scene caption using large multimodal models.}
\label{tab:caption_generation}
\end{figure*}

\section{Task Prompts}
\label{app:task_prompts}

\subsection{Single character in multiple visual contexts}
\label{app:single_multiple}

We present the prompt used to generate multi-scene text descriptions for single character in multiple visual contexts from GPT-4 in Figure \ref{tab:eval_prompt_same_char_diff_context}. 

\begin{figure*}[h]
\centering
\resizebox{\linewidth}{!}{

\begin{tabular}{p{1.3\linewidth}}

\toprule

Create four concise continuous movie scenes (7-10 words) focusing on a specific real-world character. The scenes should form a cohesive narrative.\\\\

Guidelines:\\\\

Choice of Character: Select a real-world animal as the focal point of your scenes.\\
Scene Description: Clearly describe the setting, actions, and any notable elements in each scene.\\
Connection: Ensure that the scenes are logically connected, with the second scene following on from the first.\\
Brevity and Precision: Keep descriptions short yet vividly detailed.\\\\

Example:\\\\

Character: polar bear\\
Scene 1: A polar bear navigates through a icy landscape.\\
Scene 2: The polar bear hunts seals near a crack in the ice.\\
Scene 3: The polar bear feasts on the seal.\\
Scene 4: The polar bear curls up for a nap.\\\\

Now it's your turn.\\
\bottomrule
\end{tabular} }

\caption{GPT-4 Prompt to generate multi-scene prompts for single character in multiple visual contexts.}
\label{tab:eval_prompt_same_char_diff_context}
\end{figure*}

\subsection{Different characters in a specific visual context}
\label{app:diff_specific}

We present the prompt used to generate multi-scene text descriptions for different characters in a specific visual context from GPT-4 in Figure \ref{tab:eval_prompt_diff_char_same_vis_context}.

\begin{figure*}[h]
\centering
\resizebox{\linewidth}{!}{

\begin{tabular}{p{1.3\linewidth}}

\toprule
Create four concise scene descriptions (7-10 words) where different characters perform identical action/events.\\\\

Choice of Characters: Select four real-world animals as the focal point of the individual scenes.\\
Background Consistency: Ensure that the background is consistent in both the scenes.\\
Brevity and Precision: Keep descriptions short yet vividly detailed.\\\\

Example:\\
Characters: teddy bear, panda, grizzly bear, polar bear\\
Scene 1: A teddy bear swims under water.\\
Scene 2: A panda swims under the same water.\\
Scene 3: A panda swims under the same water.\\
Scene 4: A panda swims under the same water.\\\\

Now it's your turn.\\
\bottomrule
\end{tabular} }

\caption{GPT-4 Prompt to generate multi-scene prompts for different characters in a specific visual context.}
\label{tab:eval_prompt_diff_char_same_vis_context}
\end{figure*}

\section{Automatic Evaluation Prompt}
\label{app:automatic_eval_prompt}

We present the prompt used to perform automatic evaluation of the multi-scene generated videos using large multimodal model, GPT-4-Vision, in Figure \ref{tab:auto_eval_prompt}. We utilize GPT-4-Vision for automatic evaluation since it can reason over multiple images. Specifically, we provide the multimodal model with four video frames for each scene in the generated video. The model has to provide its judgments based on the entity consistency, background consistency, and text adherence of the video frames.

\begin{figure*}[h]
\centering
\resizebox{\linewidth}{!}{

\begin{tabular}{p{1.3\linewidth}}

\toprule
You are a capable video evaluator. You will be shown a text script with two-scene descriptions where the events/actions . Video generating AI models receive this text script as input and asked to generate relevant videos. You will be provided with eight video frames from the generated video. Your task is to answer the following questions for the generated video. \\
1. Entity Consistency: Throughout the video, are entities consistent? (e.g., clothes do not change without a change described in the text script) \\ 
2. Background Consistency: Throughout the video, is the background consistent? (e.g., the room does not change described in the text script) \\
3. Text Adherence: Does the video adhere to the script? (e.g., are events/actions described in the script shown in the video accurately and in the correct temporal order) \\\\

Respond with NO, PARTIALLY, and YES for each category at the end. Do not provide any additional explanations. \\\\

Two-scene descriptions: \\\\

Scene 1: \{scene1\}\\
Scene 2: \{scene2\}\\
\bottomrule
\end{tabular} }

\caption{Prompt used to perform automatic evaluation of the multi-scene generated videos. We use this prompt when the number of scenes in the task prompt is two.}
\label{tab:auto_eval_prompt}
\end{figure*}

\section{Human Evaluation Screenshot}
\label{sec:human_ss}

\begin{figure*}[h]
    \centering
    \includegraphics[scale=0.42]{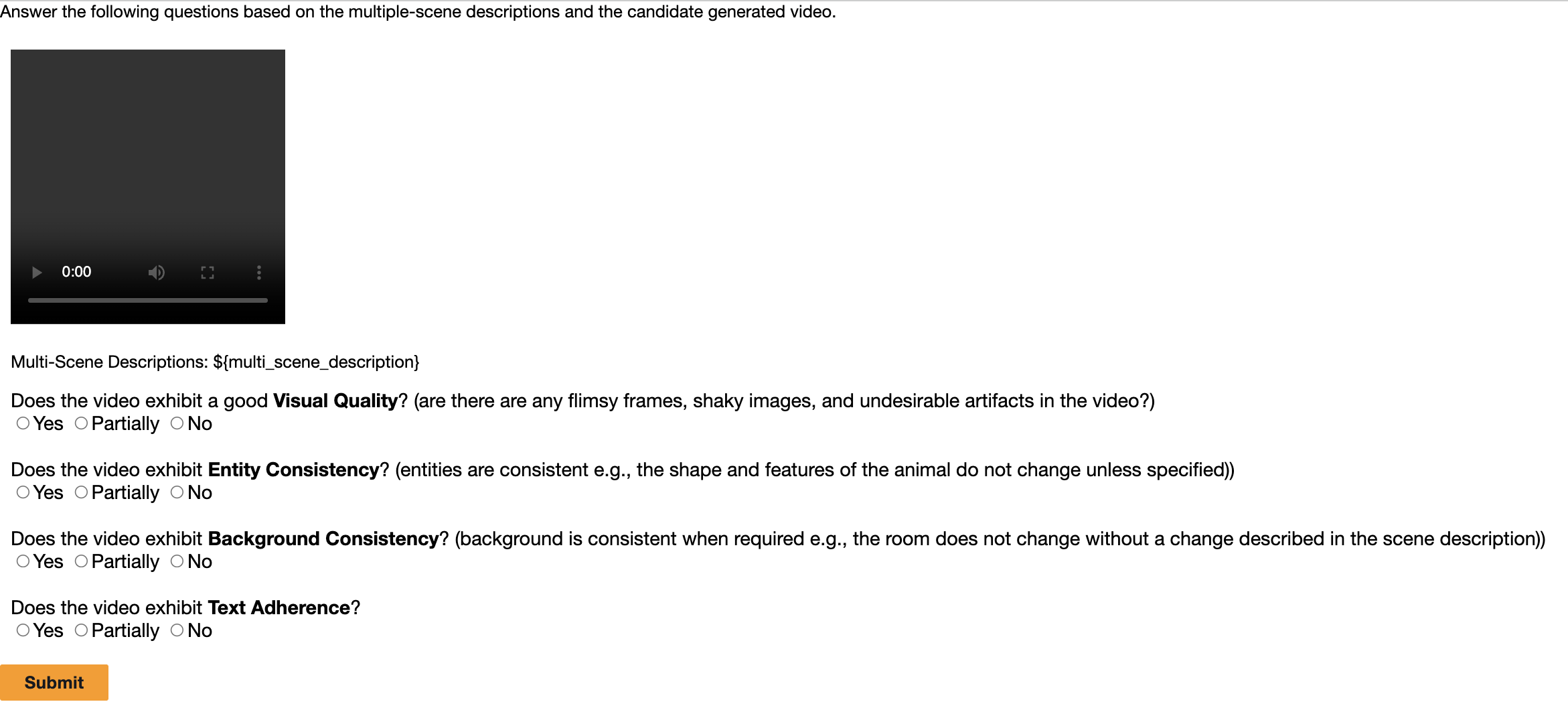}
    \caption{Human Annotation Layout}
    \label{fig:human_layout}
\end{figure*}

We present the screenshot for the human evaluation in Figure \ref{fig:human_layout}. Specifically, we ask the annotators to judge the visual quality, entity consistency, background consistency, and text adherence of the multi-scene generated videos across diverse task prompts and number of scenes.

\section{Fine-grained Analysis}
\label{sec:fine_grained_analysis}

We present the automatic evaluation results across task prompts (\S \ref{sec:fine_grained_analysis_task_prompts}) and number of scenes (\S \ref{sec:fine_grained_analysis_num_scenes}).

\subsection{Task Prompts}
\label{sec:fine_grained_analysis_task_prompts}

We compare the performance of the baselines and \name framework across different task prompts in Table \ref{tab:app_fine_grained_task_prompts}. We find that the \name-finetuned model achieves the highest overall score over all the baselines. Specifically, we find that the \name framework achieves a high visual consistency with scores close to the merging captions baseline. Further, we observe that the \name framework achieves a higher text adherence in comparison to the merging captions, with or without finetuning, across all the task prompts.

\begin{table*}[h]
\centering
\caption{\textbf{Automatic evaluation results across task prompts.} Here, S1 refers to the single character in multiple visual contexts. S2 refers to the different characters in a specific visual context. S3 refers to the multi-scene captions from real videos. We abbreviate Finetuning as F.T., Visual consistency as V.C., Text adherence as T.A.}
\label{tab:app_fine_grained_task_prompts}
% \resizebox{\linewidth}{!}{%
\begin{tabular}{cccccccccc}
\hline
Method & \multicolumn{3}{c}{S1}          & \multicolumn{3}{c}{S2}          & \multicolumn{3}{c}{S3}          \\\hline
& V.C. & T.A. & Overall & V.C. & T.A. & Overall & V.C. & T.A. & Overall \\\hline
Merge captions & 95.9                & 43.8       & 69.9    & 99.5                & 21.5       & 60.5    & 81.9                & 32.0       & 56.9    \\
Merge videos & 69.4                & 71.5       & 70.5    & 71.2                & 82.3       & 76.7    & 57.9                & 58.7       & 58.3    \\
F.T. w/ merge captions & 94.2                & 57.1       & 75.7    & 94.9                & 31.2       & 63.1    & 50.8                & 24.3       & 37.5    \\\hline
\name & 93.6                & 48.3       & 71.0    & 99.3                & 57.3       & 78.3    & 81.4                & 36.1       & 58.8    \\
F.T. w/ \name & 93.3                & 62.6       & \textbf{77.9}    & 98.9                & 76.3       & \textbf{87.6}    & 79.7                & 47.9       & \textbf{63.8}   \\\hline
\end{tabular}
%}
\end{table*}

\subsection{Number of Scenes}
\label{sec:fine_grained_analysis_num_scenes}

We compare the performance of the baselines and \name framework across different number of generated scenes in Table \ref{tab:app_fine_grained_num_scenes}. We find that the \name-finetuned model outperforms all the baselines in this setup. In addition, we find that the visual consistency of the \name framework does not change much with the number of the scenes. However, we find that the text adherence of the baselines and \name framework reduces with the number of generated scenes. The text adherence scores of the merging videos does not change with the number of scenes as it generates the videos for the individual scenes independently.

\begin{table*}[h]
\centering
\caption{\textbf{Automatic evaluation results across different number of scenes in the task prompts.} We abbreviate Finetuning as F.T.,  visual consistency as V.C., and Text Adherence as T.A.}
\label{tab:app_fine_grained_num_scenes}
% \resizebox{\linewidth}{!}{%
\begin{tabular}{cccccccccc}
\hline
                          & \multicolumn{3}{c}{\# scenes = 2} & \multicolumn{3}{c}{\# scenes = 3} & \multicolumn{3}{c}{\# scenes = 4} \\\hline
                          & V.C.        & T.A.       & Overall       & V.C.        & T.A.       & Overall       & V.C.        & T.A.       & Overall       \\\hline
Merge captions            & 93.2        & 34.4       & 63.8          & 92.5        & 33.0       & 62.7          & 87.4        & 29.9       & 58.6          \\
Merge videos             & 66.7        & 69.9       & 68.3          & 65.2        & 71.9       & 68.6          & 63.5        & 70.7       & 67.1          \\
F.T. w/ merge captions & 87.7        & 45.7       & 66.7          & 83.2        & 39.5       & 61.4          & 60.0        & 27.4       & 43.7          \\\hline
\name                      & 92.6        & 54.4       & 73.5          & 89.8        & 48.0       & 68.9          & 87.3        & 39.3       & 63.3          \\
F.T. w/ \name    & 88.5        & 66.6       & \textbf{77.5}          & 90.7        & 64.1       & \textbf{77.4}          & 87.8        & 56.1       & \textbf{71.9}  \\\hline       
\end{tabular}
%}
\end{table*}

\section{Inference Details}
\label{sec:inference_details}

We provide the details for sampling multi-scene videos from the ModelScope and Lumiere T2V models in Table \ref{tab:inference_modelscope} and Table \ref{tab:inference_lumiere}, respectively.

\begin{table*}[h]
\centering
\caption{Sampling setup for ModelScope T2V model.}
\label{tab:inference_modelscope}
\begin{tabular}{cc}
\hline
Resolution                       & 256 $\times$ 256                   \\
Number of video frames per scene & 16                          \\
Guidance scale                   & 12                          \\
Sampling steps                   & 100                         \\
Noise scheduler                        & DPMSolverMultiStepScheduler \\\hline
\end{tabular}
\end{table*}

\begin{table*}[h]
\centering
\caption{Sampling setup for Lumiere T2V model.}
\label{tab:inference_lumiere}
\begin{tabular}{cc}
\hline
Resolution                       & 1024 $\times$ 1024          \\
Number of video frames per scene & 80                          \\
Guidance scale                   & 8                          \\
Sampling steps                   & 256                         \\
Noise scheduler                  & DPMSolverMultiStepScheduler \\\hline
\end{tabular}
\end{table*}

\hb{add FreeNoise details too}

\section{Multi-Scene Data Statistics}
\label{sec:multi_scene_data_stats}

We provide the details for the multi-scene video-text dataset in Table \ref{tab:multi_scene_stats}.

\begin{table*}[h]
\centering
\caption{Multi Scene Video-Text Data Statistics}
\label{tab:multi_scene_stats}
\begin{tabular}{cc}
\hline
Number of entire videos              & 7107  \\
\% of single scene                    & 27.3\% \\
\% of two scenes                      & 25.4\%   \\
\% of three scenes                    & 31.3\% \\
\% of four scenes                     & 16.0\% \\
Number of video scene-caption instances & 20177 \\
\hline
\end{tabular}
\end{table*}

\section{Finetuning Details}
\label{sec:finetuning_details}

We provide the details for finetuning ModelScope T2V model with \name framework in Table \ref{tab:finetuning_modelscope}.

\begin{table*}[h]
\caption{Training details for the \name-finetuned ModelScope T2V model.}
\label{tab:finetuning_modelscope}
\centering
\begin{tabular}{cc}
\hline
Base Model                       & ModelScope \tablefootnote{\url{https://huggingface.co/damo-vilab/text-to-video-ms-1.7b/tree/main}} \\
Trainable Module                 & UNet                                                                                                     \\
Frozen Modules                   & Text Encoder, VAE                                                                                        \\
Batch size                       & 20                                                                                                       \\
Number of GPUs                   & 5 Nvidia A6000                                                                                           \\
Resolution                       & 256 $\times$ 256                                                                            \\
Crop                             & CenterCrop                                                                                               \\
Learning Rate Scheduler          & Constant                                                                                                 \\
Peak LR                          & 1.00E-05                                                                                                 \\
Warmup steps                     & 1000                                                                                                     \\
Optimizer                        & Adam (0.9, 0.999, 1e-2, 1e-8) \cite{kingma2014adam}                                                                           \\
Max grad norm                    & 1                                                                                                        \\
precision                        & fp16                                                                                                     \\
Noise scheduler                  & DDPM                                                                                                     \\
Number of frames per video scene & 8                                                                                                        \\
prediction type                  & epsilon     \\\hline                                                                                            
\end{tabular}
\end{table*}

\section{Single-Scene Video Generation}
\label{sec:single_scene_eval}

% \begin{figure*}[t]
% \centering
% {\includegraphics[width=0.8\textwidth]{images/time_aligned_examples.pdf}}
% \caption{{{\bf Examples of videos generated by Time-Aligned Captions (TALC)}}}
% \label{fig:time_aligned_examples}
% \end{figure*}

\begin{figure*}[t]
\centering
{\includegraphics[width=1.0\textwidth]{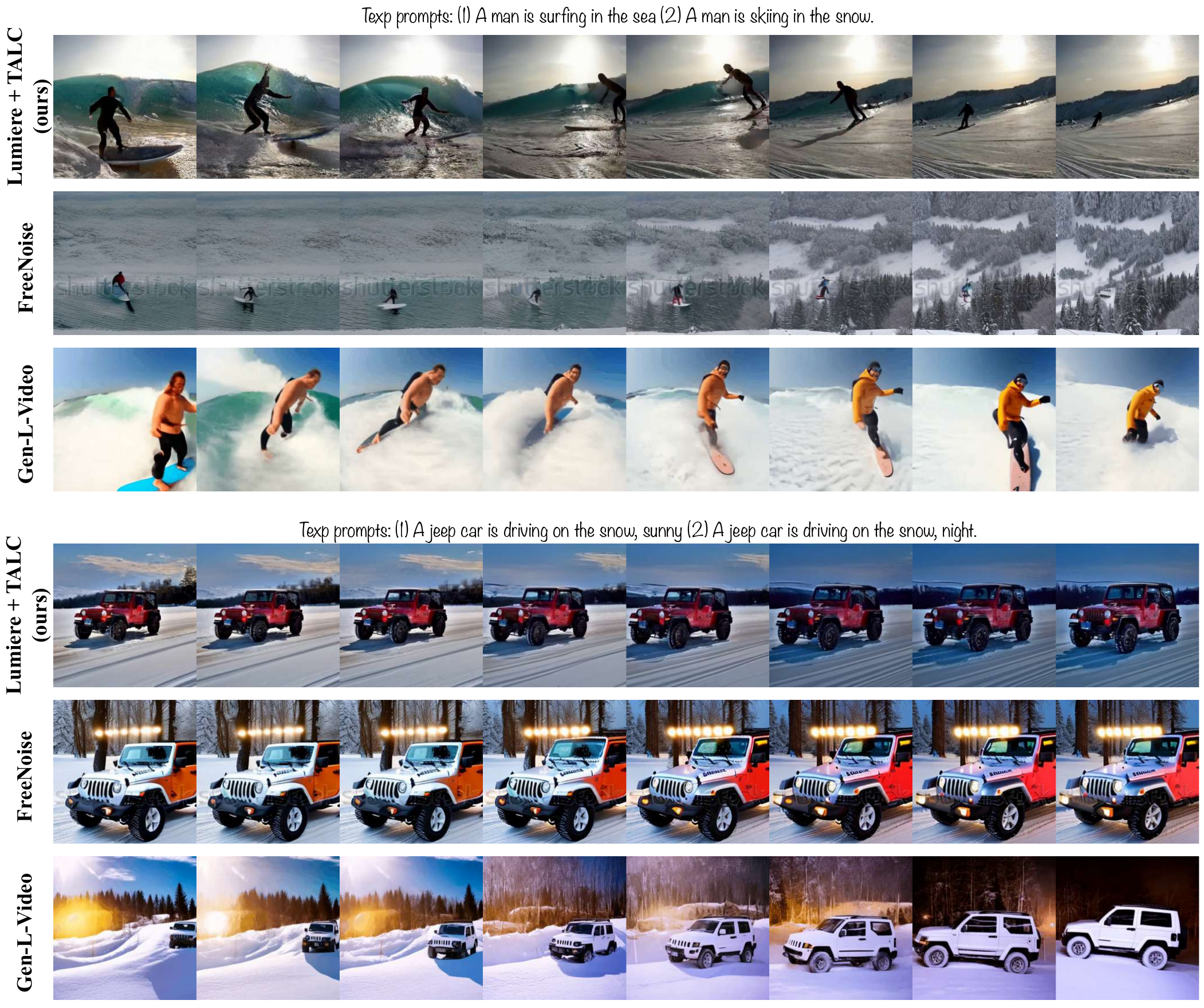}}
\caption{{\my{{\bf Examples of videos generated by TALC with Lumiere, FreeNoise and Gen-L-Video.} The videos generated using TALC with Lumiere exhibited smooth motion, seamless transitions between text prompts, and high overall quality. In contrast, videos created with FreeNoise were of lower quality, with notably less motion. For example, in the second video, the background remained static except for darkening, and the jeep did not appear to be in motion. Videos generated with Gen-L-Video demonstrated less smooth transitions between prompts and, at times, showed inconsistencies in object representation, as seen in the first video where the man's appearance changed repeatedly.}}}
\label{fig:time_aligned_examples}
\end{figure*}

\begin{figure*}[h]
    \centering
    \includegraphics[width=0.85\linewidth]{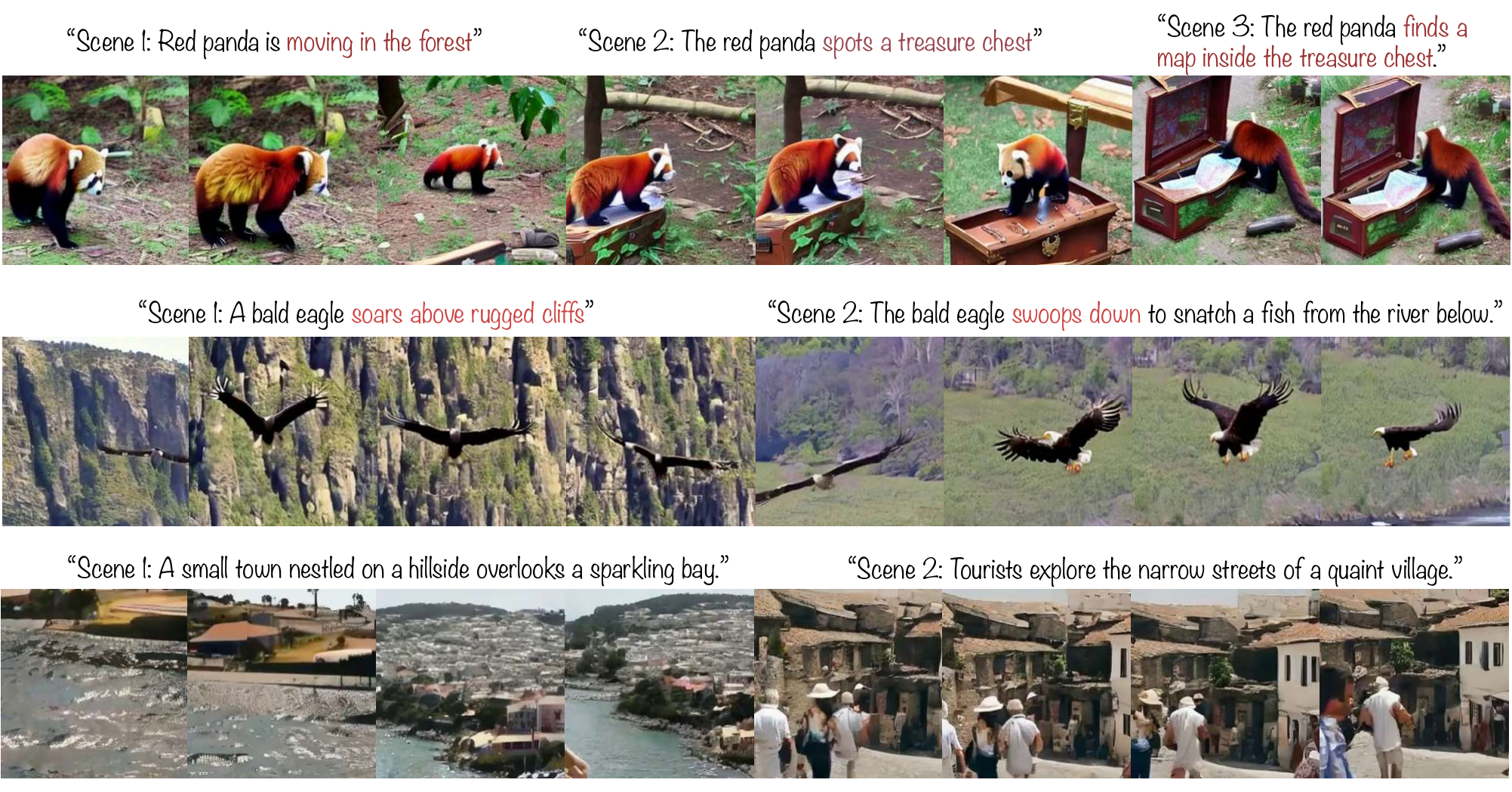}
    \caption{\textbf{Qualitative examples of videos generated by TALC with ModelScope.} Videos generated with TALC allow for the creation of videos where a single object performs multiple actions while maintaining visual consistency. Furthermore, TALC facilitates the generation of videos with diverse scenes, as demonstrated in the final example.}
    \label{fig:modelscope_ex}
\end{figure*}

To ascertain that our model's new multi-scene generation function does not detract from its single-scene generation performance, we conducted a series of evaluations using the VBench framework~\cite{huang2023vbench}. VBench offers a robust analysis of various video generation aspects such as adherence to text prompts, stylistic integrity, semantic coherence, and overall aesthetic quality.

Our analysis, shown in Table~\ref{tab:single_scene_eval}, establishes a refined baseline: ModelScope (Single-Scene Finetuning), fine-tuned on single-scene video generation data. This process yielded an average score of 0.48, indicating a decrease from the base ModelScope's average score of 0.63. This suggests that the optimizations in the base model, such as integrating high-quality images, are not fully utilized in single-scene fine-tuning.

Interestingly, fine-tuning the model on multi-scene data (ModelScope - Multi-Scene Finetuning) resulted in improved performance with an average score of 0.59, surpassing the single-scene fine-tuned version. This indicates that multi-scene data enriches the model's understanding of video content, enhancing both multi-scene and single-scene video generation.

This comparison highlights the importance of data curation and fine-tuning strategies, showing that our approach not only enables complex multi-scene narratives but also improves single-scene video generation.

% Table~\ref{tab:single_scene_eval} presents the findings, indicating that our model, despite a modest decrease in performance in some dimensions, retains most of the single-scene video quality. With a score of 0.59 against the baseline's 0.63, it retains 93.6\% of the baseline's performance, showcasing the model's robustness. The most significant variations are found in `object class' and `temporal flickering', suggesting avenues for enhancement. This retention of quality underscores the model's adaptability for various video generation tasks, confirming that the expansion into multi-scene narrative generation is achieved with a negligible impact on single-scene video quality.

% \begin{table}[ht]
% \centering
% \caption{Single-Scene Evaluation Results using VBench. \hb{Add more details. The main takeaway.}}
% \label{tab:single_scene_eval}
% \begin{tabular}{@{}lcc@{}}
% \toprule
% Dimension & ModelScope (Base) & ModelScope (\name Finetuning) \\
% \midrule
% Appearance style & 0.23 & 0.24 \\
% Color & 0.85 & 0.83 \\
% Human action & 0.96 & 0.92 \\
% Object class & 0.86 & 0.77 \\
% Overall consistency & 0.26 & 0.26 \\
% Spatial relationship & 0.35 & 0.29 \\
% Subject consistency & 0.90 & 0.83 \\
% Temporal flickering & 0.97 & 0.89 \\
% Temporal style & 0.26 & 0.25 \\ \midrule
% \textbf{Average} & 0.63 & 0.59 \\
% \bottomrule
% \end{tabular}
% \end{table}

\begin{table*}[ht]
\centering
\caption{Single-Scene Evaluation Results using VBench, comparing the base model, when it is fine-tuned on multi-scene data, and single-scene data (`f.t.' stands for fine-tuned). Our analysis shows ModelScope (Single-Scene Finetuning) as a refined baseline with an average score of 0.48, compared to the base ModelScope's 0.63. Fine-tuning on multi-scene data (ModelScope - Multi-Scene Finetuning) yields an improved score of 0.59, highlighting the efficacy of multi-scene data in enhancing video generation performance.}
\label{tab:single_scene_eval}
% \resizebox{\linewidth}{!}{%
\begin{tabular}{@{}lcccc@{}} % Add one more "c" here for the additional column
\toprule
Dimension & ModelScope (Base) & ModelScope (Multi-scene f.t.) & ModelScope (Single-Scene f.t.) \\ % Add your new column header here
\midrule
Appearance style & 0.23 & 0.24 & 0.21 \\
Color & 0.85 & 0.83 & 0.78 \\
Human action & 0.96 & 0.92 & 0.75 \\
Object class & 0.86 & 0.77 & 0.42 \\
Overall consistency & 0.26 & 0.26 & 0.22 \\
Spatial relationship & 0.35 & 0.29 & 0.14 \\
Subject consistency & 0.90 & 0.83 & 0.75 \\
Temporal flickering & 0.97 & 0.89 & 0.86 \\
Temporal style & 0.26 & 0.25 & 0.21 \\ 
\midrule
\textbf{Average} & 0.63 & 0.59 & 0.48 \\
\bottomrule
\end{tabular}%
% }
\end{table*}

\section{FreeNoise failure cases}
\label{app:freenoise_failure}

\begin{figure*}[h]
    \centering
    \includegraphics[width=0.8\linewidth]{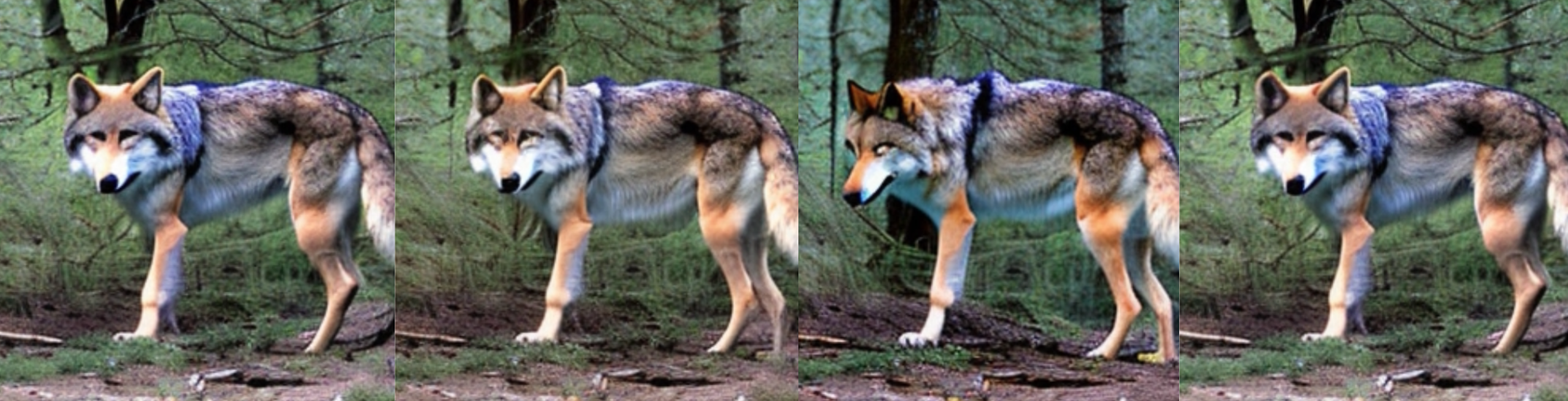}
    \caption{FreeNoise generation for the prompt: \textit{"A wolf howls at the moon in a dense forest.; The wolf prowls stealthily through the underbrush, eyes gleaming.; The wolf catches a scent and follows it eagerly.; The wolf emerges victorious, holding a fresh kill."} 
    \textbf{Explanation:} The generated scene shows a wolf standing in a forest, but the key actions described in the prompt are missing. The wolf does not prowl, its eyes are not gleaming, it does not follow any scent, and no kill is presented. This indicates a significant gap in the model's ability to convey the sequential and dynamic nature of the narrative.}
    \label{fig:freenoise_wolf}
\end{figure*}

\begin{figure*}[h]
    \centering
    \includegraphics[width=0.8\linewidth]{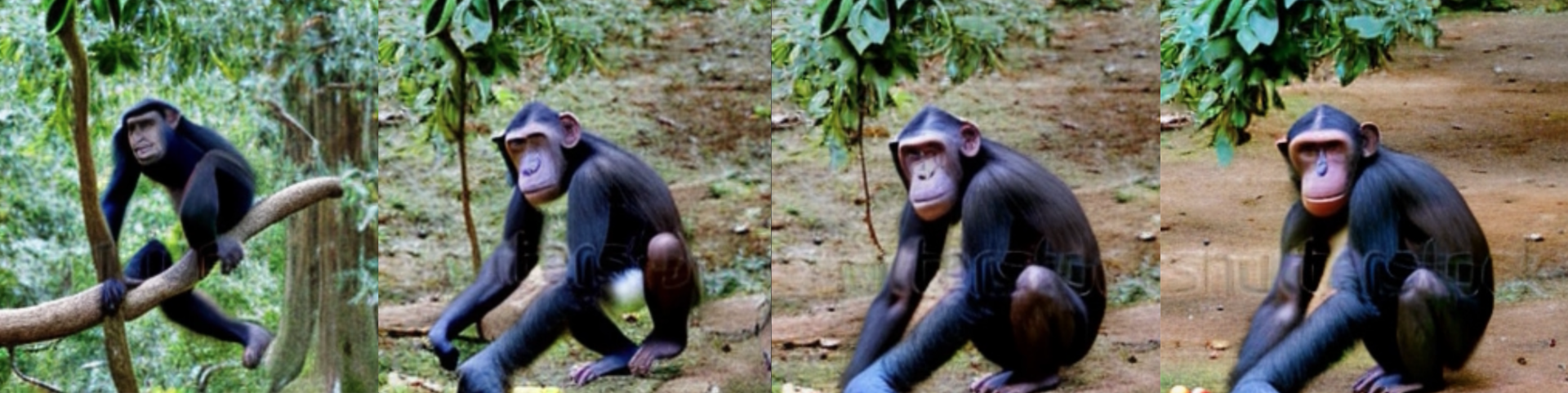}
    \caption{FreeNoise generation for the prompt: \textit{"A chimpanzee swings effortlessly through the forest canopy.; The chimpanzee gathers fruit, chattering with companions."}
    \textbf{Explanation:} Instead of depicting the chimpanzee swinging through the canopy, the generated scene shows it sitting on a tree branch and then on the ground. Furthermore, the scene lacks the critical elements of the prompt, such as the presence of fruits and companions, which are essential for conveying the intended narrative.}
    \label{fig:freenoise_chimp}
\end{figure*}

\begin{figure*}[h]
    \centering
    \includegraphics[width=0.8\linewidth]{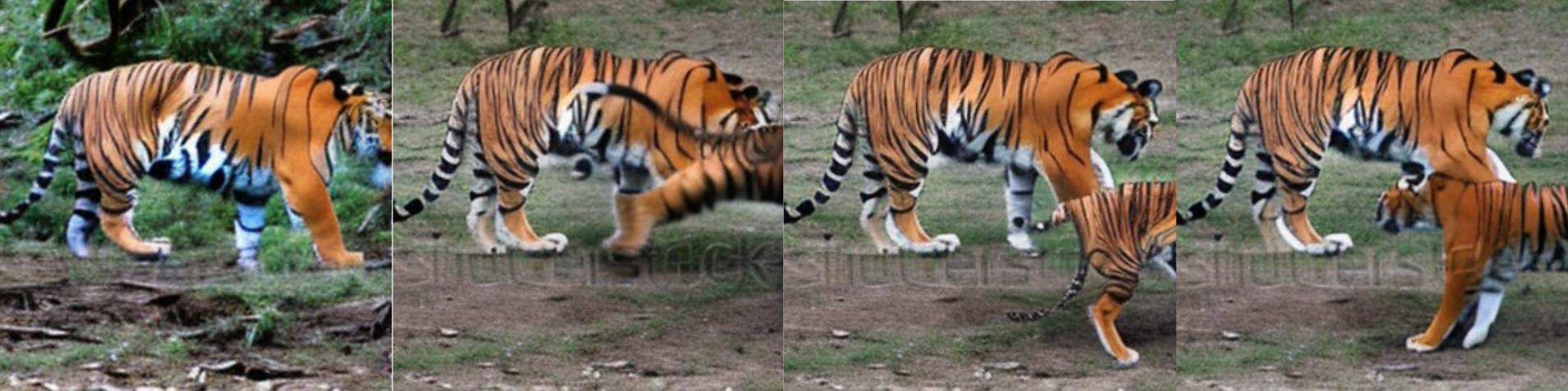}
    \caption{FreeNoise generation for the prompt: \textit{"A tiger prowls through dense jungle undergrowth stealthily.; The tiger ambushes a deer, swift and deadly."}
    \textbf{Explanation:} The scene fails to show the tiger ambushing a deer as described. Instead, the tiger is depicted standing calmly, and no deer is visible. Additionally, a second tiger appears unexpectedly from the first tiger's body, abruptly changing directions, which introduces further inconsistencies and detracts from the intended narrative.}
    \label{fig:freenoise_tiger}
\end{figure*}

\hb{In this work, we observe that the state-of-the-art method FreeNoise \cite{qiu2023freenoise} does not perform well on our multi-scene prompts. We provide some qualitative examples to show its inability to introduce new content and limited motions as multiple scenes are requested in Figure \ref{fig:freenoise_wolf}, \ref{fig:freenoise_chimp}, and \ref{fig:freenoise_tiger}}.

\end{document}